\RequirePackage{fix-cm}
\documentclass[smallextended]{for_arxiv}    

\smartqed  
\usepackage{graphicx}
\usepackage{natbib}
\usepackage{amsmath}
\usepackage{amssymb}
\usepackage{bbm}
\usepackage{rotating}
\usepackage{color}

\def\pmb#1{\setbox0=\hbox{#1}%
  \kern-.02em\copy0\kern-\wd0
  \kern-.035em\copy0\kern-\wd0
  \kern-.02em\raise.03em\box0 }
  
\newcommand{\matx}[1]{\mbox{\bf $#1$}}
\newcommand{\vect}[1]{\mbox{{\boldmath $#1$}}}

\newcommand{\mm}[1]{\mbox{\bf $#1$}}
\newcommand{\vv}[1]{\mbox{{\boldmath $#1$}}}
\newcommand{\cc}[1]{{\cal{#1}}}

\newcommand{\tp}{^{\top}}

\def\p3{\cc{P}^3}
\def\avect{\mbox{\boldmath $a$}}

\def\wvect{\mbox{\boldmath $w$}}
\def\xvect{\mbox{\boldmath $x$}}
\def\yvect{\mbox{\boldmath $y$}}
\def\zvect{\mbox{\boldmath $z$}}

\def\Mvect{\mbox{\boldmath $M$}}
\def\Nvect{\mbox{\boldmath $N$}}

\def\Amat{\mbox{\bf A}}

\def\Cmat{\mbox{\bf C}}

\def\Pmat{\mbox{\bf P}}
\def\Qmat{\mbox{\bf Q}}

\def\Xmat{\mbox{\bf X}}

\def\Xbarre{\overline{\Xmat}}
\def\wbarre{\overline{\wvect}}

\newcommand{\argmin}{\operatornamewithlimits{argmin}}

\def\DTW{\mbox{DTW}}
\def\DFW{\mbox{DFW}}
\usepackage{booktabs}
\usepackage{url}
\usepackage{latexsym}
\usepackage{float}
\usepackage{array}
\usepackage[pagebackref=true,breaklinks=true,letterpaper=true,colorlinks,bookmarks=false]{hyperref}

\usepackage[final]{review} 
\setrevision{1}

\begin{document}

\title{Continuous Action Recognition Based on Sequence Alignment\thanks{The authors acknowledge support from the European project HUMAVIPS
                  \#247525 (2010-2013) and from the ERC Advanced Grant VHIA \#340113 (2014-2019). J. Cech acknowledges support from the Czech Science Foundation Project GACR P103/12/G084.}}

\titlerunning{Continuous Action Recognition Based on Sequence Alignment}

\author{Kaustubh Kulkarni \and Georgios Evangelidis \and \\ Jan Cech \and Radu Horaud}
\authorrunning{K. Kulkarni, G. Evangelidis, J. Cech \& R. Horaud}

\institute{
Kaustubh Kulkarni \at
INRIA Grenoble Rh\^one-Alpes\\
Montbonnot Saint-Martin,  FRANCE\\
\email{Kaustubh.Kulkarni@inria.fr}
\and
Georgios Evangelidis \at
INRIA Grenoble Rh\^one-Alpes\\
Montbonnot Saint-Martin,  FRANCE\\
\email{Georgios.Evangelidis@inria.fr}
\and
Jan Cech \at
Center for Machine Perception\\
Czech Technical University in Prague, Czech Republic\\
\email{cechj@cmp.felk.cvut.cz}
\and
Radu Horaud \at
INRIA Grenoble Rh\^one-Alpes\\
Montbonnot Saint-Martin,  FRANCE\\
\email{Radu.Horaud@inria.fr}
}

\maketitle

\begin{abstract}
Continuous action recognition is more challenging
than isolated recognition because classification and segmentation must be simultaneously carried out. We build on the well known dynamic time warping (DTW) framework and devise a novel visual alignment technique, namely \textit{dynamic frame warping} (DFW), which performs isolated recognition based on per-frame representation of videos, and on aligning a test sequence with a model sequence. Moreover, we propose two extensions which enable to perform recognition concomitant with segmentation, namely one-pass DFW and two-pass DFW. These two methods have their roots in the domain of continuous recognition of speech and, to the best of our knowledge, their extension to continuous visual action recognition has been overlooked. We test and illustrate the proposed techniques with a recently released dataset (RAVEL) and with two public-domain datasets widely used in action recognition (Hollywood-1 and Hollywood-2). We also compare the performances of the proposed isolated and continuous recognition algorithms with several recently published methods. 
\end{abstract}


\section{Introduction}
\label{section:introduction}

The problem of action recognition
from temporal visual information
is a very active research topic with many challenging applications. A large majority of existing action recognition techniques assumes that the boundaries (first and last frames) of individual actions are known in advance. This gives rise to the per-video approach, i.e., isolated action recognition, where an action label is assigned to a whole video to be recognized, consequently the latter is described by a single vector, e.g., the bag-of-word histogram representation framework. This whole-video paradigm allows one to solve isolated recognition problems using discriminative classifiers. In this paper we address the more realistic \textit{continuous action recognition} problem, i.e., a video may contain a sequence of unknown actions in an unknown order and with unknown boundary locations between consecutive actions. This continuous recognition problem is more difficult than the isolated recognition problem because one has to address both classification and segmentation. The discriminative isolated-recognition framework that we just mentioned cannot be easily generalized to deal with segmentation.

In this work, continuous recognition is addressed in the framework of dynamic time warping (DTW) which has the potential to handle action-level classification and sequence-level segmentation  in a concomitant and consistent manner. The proposed methodology has the following original components: a per-frame time-series representation of videos, a template-based representation of action categories, and a template-to-data alignment process that assigns a label to each video frame. The method will be referred to as \textit{dynamic frame warping} (DFW) and two DFW implementations are proposed, namely one-pass (OP-DFW) and  two-pass (TP-DFW) dynamic frame warping. Both these dynamic programming (DP) algorithms have their roots in the speech domain but, to the best of our knowledge, the extension of continuous speech recognizers to visual action recognition has been overlooked. 

In speech, the DP-based sequence-to-sequence alignment framework gave rise to two extensions in order to deal with the problem of word recognition from a continuous speech signal, namely one-pass \cite{vintsyuk1971element} and two-pass \cite{Sakoe1979} algorithms.
The one-pass DP approach is used in conjunction with either dynamic time warping (DTW) 
\cite{ney1984use}, \cite{NeyOrtmanns1999} 
or with the Viterbi algorithm
\cite{lee1989frame}, and it is used today by large-vocabulary continuous speech recognition systems \cite{gales2008application}.  
The one-pass hidden Markov model (HMM), adapted from speech, has been used by continuous sign-language recognizers 
\cite{starner1998real}, \cite{vogler1998asl}. 

The potential attractiveness of the two-pass DP algorithm is that its first pass (action-level) can be carried out with virtually \textit{any} (generative or discriminative) isolated-action recognition method.
 It is worth noticing that a few recent continuous action recognition methods use a two-pass strategy in combination with HMMs and AdaBoost \cite{LvNevatia2006}, with SVM and semi-Markov models (SMM) \cite{ShiWanCheSmoijcv11}, and with multi-class SVM  \cite{HoaiLD11}. 

Sequence alignment algorithms can use either tem\-plate-based methods, e.g., DTW, or probabilistic methods, e.g., hidden Markov models (HMMs). Within the context of action recognition, an HMM must be associated with each action category. This means that one needs to define a set of states for each category and to estimate the HMM parameters for each category, namely the state-transition probabilities and the state-emission probabilities. These action-level HMMs require a large amount of training examples. Moreover, continuous re\-cog\-nition requires that between-action (or \textit{jump}) probabilities are estimated from an annotated set of videos containing a large number of action-jump examples. Altogether, HMM-based action-level continuous recognition needs a huge amount of training data which may not be available in practice.

One way to reduce the complexity of HMMs is to define action sub-units such that a large number of action types can be described by a small catalog of sub-units. For example, speech recognizers have successfully used word sub-units such as phonemes, syllables, etc. In the speech domain these sub-units have a clear acoustic and linguistic interpretation. Indeed, there are well defined and well understood language-dependent rules allowing the concatenation of pho\-nemes into words from which within-word state-transition probabilities can be estimated. Moreover, there are grammar rules from which be\-tween-word transitions can be inferred. Unfortunately, the definition and use of a small catalog of action sub-units turns out to be problematic even in restricted hand-gesture domains, e.g., sign-language recognition \cite{starner1998real,vogler1998asl,liang1998real,HienzBauerKraiss1999,vogler2001framework}. More generally, the use of HMMs for the problem of continuous action recognition would require huge amounts of annotated videos in order to train action sub-unit HMMs (one HMM per sub-unit) as well as action-level and sequence-level HMMs. 

Another drawback of both DP and HMM continuous recognition approaches is that their performance crucially depends upon correctly labeling the first frame of an action to be recognized. This happens because current models cannot account for temporal dependencies beyond the first- or second-order Markov assumptions. The immediate consequence of mislabeling the first frame is that the DP procedure propagates this error to the next frames. One solution is to use a hidden conditional random fields (HCRF) which can deal with long temporal-range data dependencies: Rather than depending on the current state, as in HMM, the data are conditionally dependent of all the HCRF's states. Nevertheless, HCRF suffer from the same difficulty of properly inferring between-action transition probabilities from the training data. 

In order to circumvent the above mentioned pro\-blems encountered with current approaches, we propose a novel way of representing actions based on \textit{action templates} and we show how this representation can be cast in one-pass and two-pass DP algorithms.
We address the problem of devising a template representation that captures the inherent actor-dependent variabilities occurring within action categories. We also address the problem of how to encode \textit{jumps} between templates, such that the first and last frames of every action, present in a long sequence of unknown actions, are robustly detected and labeled. To summarize, the paper has the following contributions:

\begin{figure*}[t!]
\begin{center}
\begin{tabular}{cc}
\includegraphics[width=0.38\linewidth]{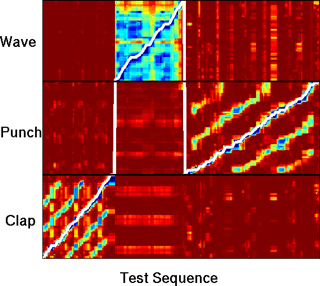}
\includegraphics[width=0.38\linewidth]{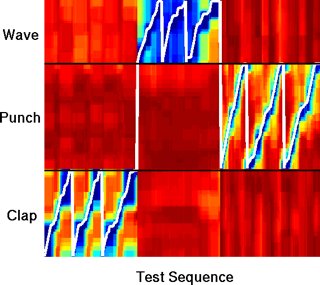}\\
\end{tabular}
\bigskip
\begin{tabular}{c|c}
\includegraphics[width=0.48\textwidth]{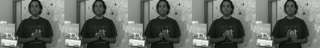} &
\includegraphics[width=0.48\textwidth]{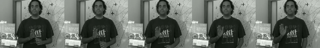} \\
{\fontsize{8}{8}\selectfont Clap} & {\fontsize{8}{8}\selectfont Wave}
\end{tabular}
\end{center}
\caption{
The proposed one-pass dynamic frame warping algorithm performs simultaneous action recognition and segmentation. Like any other DTW-based method, our algorithm proceeds in two steps. The forward step computes an accumulated cost at each grid point (warm colors indicate high cost values) based on a test-frame-to-model-frame distance. The cost at every grid point contains within-action and between-action transition information. The backward step extracts a minimum-cost path (shown in white) that assigns an action label to each frame of the test video. The example shows three periodic actions: punch, clap and wave. Each one of these actions is composed of an arbitrary number of motion patterns. The proposed formulation offers two possibilities for dealing with periodic actions: either each periodic-action model is aligned with a test sub-sequence (top left of the figure) or each motion-pattern model is repeatedly aligned with a test sub-sequence (top right). While the former jumps from one action to the next, e.g., ``clap-wave-punch" (from the left to the right), the latter jumps from one motion pattern to the next, e.g., ``clap-clap-clap", ``wave-wave-wave", and ``punch-punch-punch" (see section \ref{sec:periodic-actions} for more details). The bottom row of the figure shows a between-action boundary detected by the algorithm.
\color{black}
}
\label{fig:continuous_rec}
\end{figure*}
\begin{itemize}
\item
Each action category is described by an \textit{ac\-tion-tem\-plate} which is a sequence that corresponds to a \textit{mean action},
optimally estimated from pairwise alignments between all the training examples of that action, and which contains \textit{within-category variability information} as well as the minimum and maximum durations over all the examples. More precisely, an \textit{action-template} is a sequence of \textit{metaframes} where a metaframe is a collection of aligned frames, thus accounting for \textit{between-frame variabilities}. As it will be described in detail below, continuous recognition also requires a structured representation of all the action categories, which is modeled as a \textit{super-template}, or a string formed by all the action templates taken in an arbitrary order. Action templates, metaframes, and super-templates are formally defined in section~\ref{section:notations_definitions}.
\item
This formulation immediately calls for a test-frame-to-template-metaframe dissimilarity function that must be plugged into the DP grid of alignment scores. The inherent frame variability, i.e., within a meta\-frame, suggests that a naive implementation, such as the minimum over the test-frame-to-template-frame Euclidean distances, is unreliable and non discriminant. We propose to compute a frame-to-metaframe dissimilarity score based on a sparse representation through solving a basis pursuit denoising problem \cite{Chen2001} that can be cast into a computationally efficient convex quadratic programming procedure.
\item
Each video frame is modeled with a high-dimensional vector that must contain enough information to characterize an action sub-unit. One must however consider a time-window centered at each frame and which is then shifted over the frames, in order to gather spatiotemporal features. There must be a minimum number of such features to properly characterize an action, regardless of the action type and of the speed at which various actors perform actions. Therefore, the proposed per-frame feature vector uses a temporal window of an adjustable size, 
such that a predefined minimum number of features is included in the frame descriptor. Moreover, we stress the fact that it is important to take into account between-action information that is encoded in the data. This is done by annotating long videos that contain action sequences and not isolated actions. Hence, the feature vector associated with the first and last frames of an action necessarily contain information allowing the recognition algorithm to detect action boundaries and hence to jump from one category to another category.
\item 
The proposed OP-DFW may well be viewed as a generalization of sequence alignment methods for the isolated-action case, which perform recognition by estimating a model-sequence-to-test-sequence alignment through evaluating a dynamic programming score. In isolated recognition the alignment is based solely on the detection of \textit{within-action transitions}. To cope with continuous recognition, OP-DFW allows \textit{between-action transitions} as well, such that it is possible to \textit{jump} from the last frame of an action to the first frame of any other action, e.g., Figure~\ref{fig:continuous_rec}. 
We show that this continuous recognition framework is particularly well suited for modeling periodic actions because it allows jumps between repetitive motion patterns, i.e., section~\ref{sec:periodic-actions}.
The ability of the algorithm to handle both these two types of transitions is one of the major contributions of this paper.
\item
The basic idea of TP-DFW, inspired from \cite{Sakoe1979}, is to carry out action-level recognition and sequence-level segmentation in two consecutive passes. Firstly, all possible sub-sequences of the unknown video are considered. Each such sub-sequence is parameterized by its first and last frames in the video. An isolated recognition algorithm is repeatedly applied to each sub-sequence, thus associating an action category and a recognition score to each sub-sequence of the initial video, namely to each possible first-frame/last-frame point on a  two-dimensional grid. Secondly, a variant of DP is used to estimate an optimal path of recognition scores through this grid, under the constraint that the actions contained in the unknown video must form a temporal string. 
\end{itemize}

Not surprisingly, one-pass DFW performs better than both the proposed two-pass DFW and the two-pass DP methods recently proposed by \cite{ShiWanCheSmoijcv11,HoaiLD11}. This can be easily explained by the fact that the training data used by our method uses annotated video examples composed of action sequences and not of isolated actions. In this way, the feature descriptors outlined above encode both within-action and between-action transitions. The proposed one-pass algorithm explicitly takes into account these two transition types, while two-pass algorithms cannot enforce between-action transitions. Indeed, the second pass, i.e., segmentation based on dynamic programming, is strong\-ly biased by the isolated recognition scores found by the first pass. 
It is worthwhile to notice that two state-of-the-art continuous action recognition methods \cite{ShiWanCheSmoijcv11,HoaiLD11} use a two-pass dynamic 
programming strategy, without describing it in detail and 
without referring to the speech recognition literature. The inclusion of a detailed description of our proposed two-pass method allows computer vision practitioners 
to easily implement this type of DP algorithms and to understand why the proposed one-pass method performs slightly better than two-pass methods. 

The remainder of this paper is organized as follows. Section~\ref{section:related_work} describes in detail the work related to our method. Section~\ref{section:notations_definitions} formally states the problem of continuous action recognition. Section~\ref{section:baseline_dtw} outlines the dynamic time warping baseline method for aligning two time-series. Section~\ref{section:class_templates} describes the proposed template-based representation of actions based on metaframes, as well as the concept of action strings. Section~\ref{section:dfw} describes the proposed dynamic frame warping algorithm based on a novel frame-to-metaframe distance. Section~\ref{section:continuous_recognition} describes in detail the one-pass and the two-pass dynamic frame warping methods, explains how periodic actions are handled, analyses the complexity of the proposed algorithms, and suggests a null-class model. Section~\ref{section:experiments} describes in detail the proposed bag-of-words representation, shows the results obtained with several datasets, and compares our algorithms with both isolated and continuous recognition methods. Section~\ref{section:conclusions} draws a few conclusions. 

Matlab code and additional multimedia material are made publicly available.\footnote{
\url{http://perception.inrialpes.fr/people/Kulkarni/IJCVMaterials/}
}
\color{black}
\section{Related Work}
\label{section:related_work}

In computer vision, the first attempts to solve for continuous action recognition addressed the problem of sign language recognition. The similarity between spoken and sign languages enabled the use of HMMs. \cite{starner1998real} considered a forty word lexicon from the American sign language (ASL) and used the hidden Markov model toolkit (HTK) \cite{young1993htk} both for inference and recognition. The difficulty of defining sub-units of signs, e.g., by similarity to phonemes in speech, lead \cite{HienzBauerKraiss1999}
to model each sign with one HMM. In order to optimally find sign boundaries the authors proposed to use tree-search for pruning too long or too short paths found by a Viterbi algorithm and to extract the between-sign boundaries. Inclusion of a stochastic language model slightly improved recognition using a lexicon of 52 signs representing 7 different word types (nouns, verbs, etc.) from the German sign language. 

Unlike spoken languages where the acoustic data are sequential, sign languages use the two hands simultaneously, and both the hands' shape and orientation occur in parallel. In order to model these features, \cite{vogler2001framework} proposed to use 200 parallel and independent HMMs and adapted the token passing algorithm \cite{young1989token} to recognize simultaneous aspects of ASL. The authors note, however, that the decomposition of language signs into pho\-nemes is controversial from a linguistic point of view. In summary, the critical components of HMM-based continuous sign language recognition are proper definitions of sign subunits and of language models.

The idea of parallelized HMMs was also explored by \cite{LvNevatia2006} in the framework of segmentation and recognition of human actions from 3-D data. The authors hypothesized that a complex action may be viewed as the combination of single-, two-, and three-joint actions. Hence they proposed to decompose complex human actions using seven such primitive actions and one HMM is associated with each such action.  A multi-class AdaBoost classifier is fed by these weak HMM classifiers. Action recognition and segmentation are implemented using the two-pass strategy already described. The same authors subsequently proposed a different HMM continuous recognition framework that associates a one-state HMM to each action category \cite{LvNevatia2007}. Hence, segmentation and recognition reduce to the problem of deciding whether to stay in the current state or to jump to another state. A uniform distribution is chosen for the transition probabilities, hence they are neglected. This model may well be viewed as a very simple variant of one-pass HMM methods used in speech. Both \cite{LvNevatia2006,LvNevatia2007} use 3-D motion capture data (using a multiple-camera setup) for model inference. While \cite{LvNevatia2006} uses the same kind of 3-D data for recognition and segmentation, \cite{LvNevatia2007} only needs a silhouette extracted from a single view, and hence the latter is more flexible than the latter. Nevertheless, proper estimation of the between-action transition probabilities severely limits the performance. 

The HMM-based generative models that we just discussed make strict assumptions that observations are conditionally independent, given class labels, and cannot describe long-range dependencies of the observations. This limitation makes the implementation of one-pass dynamic programming methods unreliable because it is difficult to decide which type of transition (within-action or between-action) should be preferred along the DP forward pass. Conditional random fields (CRFs) are discriminative models that explicitly allow transition probabilities to depend on past, present, and future observations. CRF models applied to isolated activity recognition outperform HMMs, e.g.,  \cite{SminchicescuKanaujiaMetaxas2006,vail2007conditional}. Several authors extended the CRF framework to incorporate additional latent (or hidden) state variables in order to better deal with the complex structure of human actions and gestures. For example \cite{morency2007latent} proposed a latent-dynamic CRF model, or LDCRF, to better capture both the sub-gesture and between-gesture dynamics. The method was applied to segment and classify head movements and eye gazing in a human-avatar interactive task. 

The methods described so far use motion (or pose) parameters which are extracted using motion capture systems. The characterization of actions using such parameters is attractive both because they are highly discriminant and because they live in a low-dimensional space, hence they can be easily plugged in the HMM and CRF frameworks. However, it is not always possible to reliably extract discriminant motion or pose descriptors from visual data, and sophisticated multiple-camera setups are required both for training and recognition.  Alternatively, image-based descriptors are easy to extract but the corresponding feature vectors are less discriminant and have dimensions as high as hundreds, which make them unsuitable for training graphical models. Recently \cite{ning2008latent} proposed to plug a latent pose estimator into the LDCRF model of \cite{morency2007latent} by jointly training an image-to-pose regressor and a hidden-state conditional random field model. Although appealing, this model also requires a large training set gathered with synchronized multiple-camera and motion capture systems \cite{sigal2010humaneva}.

The proposed one-pass continuous recognition algorithm also differs from recently proposed dynamic time warping methods. \cite{AlonAthistsosYuanSclaroff2009}  address the problem of continuous hand gesture recognition and a pruning strategy is proposed such that DTW paths that do not correspond to valid (trained) gestures are abandoned. At runtime, this is less efficient than one-pass DP algorithms which extract a single path rather than multiple paths. \cite{kulkarniECCV2008} address the problem of continuous action recognition but propose an average-template representation for an action category and a dissimilarity measure which would not be able to handle large intra-class variance. Dynamic time warping has also been applied to action recognition in combination with unsupervised manifold learning techniques, e.g., \cite{blackburn2007human,ZhouT09,GongMedioni2011}, but the problem of continuous recognition was not addressed in these papers.   To the best of our knowledge, the full potential of dynamic time warping for the problem of simultaneous segmentation and recognition of human actions has not been systematically exploited in the computer vision domain. 

\section{Problem Formulation and Notations}
\label{section:notations_definitions}

Let image sequences (videos) be represented as time-series,  or sequences of vectors, denoted by $X$, $Y$, or $Z$, e.g., $X_{1:T_X}=(\xvect_1,\hdots \xvect_t,\hdots \xvect_{T_X})$,  where $\xvect \in\mathbbm{R}^{K}$ is an \textit{$\ell_2$-normalized vector}, $\|\cdot\|_2=1$, that describes the $t$-th image (or frame) of a video, and $T_X$ denotes  the number of frames in $X$. The notation $X_{t_i:t_j}$ refers to a sub-sequence of $X$ starting at frame $t_i$ and ending at $t_j$, with 
$1\leq t_i < t_j \leq T_X$. 
We denote by $d (\xvect, \xvect')$ the distance between the a frame of $X$ and a frame of $X'$.

We assume that there are $L$ possible action categories, i.e., the action-label set is $\mathcal{L}=\{l\}_{l=1}^{L}$. Let $X^l$ denote a training example of action $l\in\mathcal{L}$ and let $\{X_n^l\}_{n=1}^{N_l}$ denote the set of $N_l$ single-action training examples of category $l$. A \textit{test} sequence $Z$ may contain an unknown number of actions $J$, with an unknown label sequence $M=(m_1,\hdots m_j,\hdots m_J), m_j\in\mathcal{L}$, with unknown boundaries between two consecutive actions, and in an unknown order. The problem of continuous action recognition consists in simultaneously finding a partition of $Z$ into single actions and to label each one of these actions.

We now define a representation of an action category. For each category $l$ we define an \textit{action-template} denoted by $Y^l$ and a \textit{class-template} denoted by $\tilde{Y}^l$. The action template is the \textit{center} of the single-action examples of a class, $\{X_n^l\}_{n=1}^{N_l}$, and hence it is a sequence itself. A class template is a sequence of \textit{metaframes}. A metaframe is a collection of frames from the training examples resulting from pairwise alignments $X_n^l\leftrightarrow X_k^l$ within the same class $l$. Section \ref{section:class_templates} below will describe in detail how action templates are computed, based on dynamic time warping, and how class templates are built. We also introduce a \textit{super-template}, or a string of class-templates corresponding to the $L$ categories taken in an arbitrary, yet fixed, order:
\begin{equation}
\label{eq:super-template}
\tilde{Y}^{1:L}= (\tilde{Y}^{1},\hdots \tilde{Y}^{l},\hdots \tilde{Y}^{L}).
\end{equation}
The task of continuous action recognition is to partition an unknown sequence $Z_{1:T_Z}$ into a string of $J$ sub-sequences:
\begin{equation}
\label{eq:Z-segmentation}
 \left\{ 
 \begin{array}{ll}
 Z= ( Z_{t_0:t_1}, \hdots Z_{t_{j-1}:t_j}, \hdots Z_{t_{J-1}:t_J} ) \\
 1=t_0 < t_1 \leq \hdots t_{j-1} <  t_j \leq \hdots  t_{J-1} < t_J = T_Z
 \end{array}
 \right.
 \end{equation} 
such that the sub-sequence-string  $Z$ is optimally aligned with a \textit{synthesized} string of $J$ class templates
 \[ \tilde{Y}^{m_1:m_J}= (\tilde{Y}^{m_1},\hdots \tilde{Y}^{m_j},\hdots \tilde{Y}^{m_J}), \; m_1, \hdots m_J \in \mathcal{L}.
 \]
 This can be written as the following optimization problem:
 \begin{equation}
 \label{eq:optimization-general}
 A^{\ast} (Z,\tilde{Y}^{m_1:m_J}) = \argmin_{\{\tilde{Y}^{1:L}\}} [ f (Z, \tilde{Y}^{1:L}) ],
 \end{equation}
 where $A^{\ast}$ is the optimal \textit{alignment path} over 
 \[ \{A(Z,\tilde{Y}^{m_1:m_J})\}_{m_1,m_J\in\mathcal{L}},\]
  i.e., the set of all possible alignments between the test sequence $Z$ and synthesized strings $\tilde{Y}^{m_1:m_J}$, and $f$ is a dissimilarity function between the test sequence and a string of templates. Notice that this optimization problem is not trivial because one has to consider the set $\{\tilde{Y}^{1:L}\}$ of all possible concatenations of class templates, align each one of these concatenations with the unknown sequence, and select the alignment that satisfies both an action-level optimality criterion as well as a sequence-level cirterion. Indeed, the cost function must be repeatedly evaluated by permuting the templates and changing the number of templates in the synthesized string. As it will be described in detail below, this simultaneous action recognition and segmentation problem can be robustly and efficiently handled within the framework of dynamic programming. 
 The complexity of the proposed algorithms is described in detail as well.
 \color{black}

\section{Dynamic Time Warping}
\label{section:baseline_dtw}

In this section we briefly describe the baseline DTW algorithm, e.g., \cite{RabinerJuang,Mueller2007}, for optimally aligning two sequences
$Z_{1:T_Z}$ and $Y_{1:T_Y}$ of 
 \textit{different} lengths and for estimating a dissimilarity statistics between them. The alignment is described by a \textit{path} or, more precisely, by a set of frame-to-frame assignments:
\begin{equation}
\label{eq:assignment-def}
A(Z,Y) = \{(t_1,t_1'), (t_i,t_i'),\dots, (t_{|A|},t_{|A|}') \},
\end{equation}
with $1 \leq t_i \leq T_Z$ and $1 \leq t_i' \leq T_{Y}$ being the frame indexes of the sequences $Z$ and $Y$. 
For example let $Z_{1:5}=(\zvect_t)$ and $Y_{1:4}=(\yvect_{t'})$. One possible path is given by 
\[A=\{(1,1),(2,1),(3,2),(3,3),(4,4),(5,4)\}.\]
From the general alignment formulation (\ref{eq:optimization-general}) one can derive a simpler constrained minimization problem to find 
an optimal path $A^\ast$, namely:
\begin{equation}
\label{eq:dtw}
\left\{
\begin{array}{l}
A^\ast(Z,Y) = \argmin_A  
\sum_{i=1}^{|A|} d(\zvect_{t_i}, \yvect_{t_i'}) \\
\begin{array}{ll}
\text{s.t. } &  \left\{ \begin{array}{l} (t_1,t_1') =  (1,1) \\ 
 (t_{|A|},t_{|A|}') = (T_Z,T_{Y}). \end{array} \right.
 \end{array}
 \end{array}
 \right. 
\end{equation}
This
estimates the best choice of the sum over frame-to-frame distances $d(\zvect_{t_i}, \yvect_{t_i'})$. The result of (\ref{eq:dtw}) is an optimal path $A^\ast (Z,Y) = \{(t_i^{\ast},t_i'{^{\ast}})\}_{i=1}^{|A^{\ast}|}$ as well as a dissimilarity score $\DTW (Z,Y)$ between the two time series, namely the normalized minimum yielded by (\ref{eq:dtw}). It corresponds to the normalized cost accumulated along the path, starting at $(1,1)$ and up to $(T_Z,T_Y)$ by traversing along the optimal path $A^\ast$ and is given by: 
\begin{equation}
\DTW (Z,Y) = \frac{1}{|A^\ast|} \sum_{i=1}^{|A^\ast|} d (\zvect_{t_i^\ast}, \yvect_{t_i'^\ast}).
\label{eq:dtw_distance}
\end{equation}
Although this is not a true distance because  the inequality property does not hold in general, it can be viewed as a dissimilarity statistic between $Z$ and $Y$. 
Dynamic time warping proceeds in two passes: a forward pass and a backward pass. 

During the forward pass, an \textit{accumulated cost} $D(t,t')$ and a \textit{back-pointer} $\Phi (t,t')$ are estimated at each grid point $(t,t')$, with $t\in\{2 \dots T_Z\}$, $t'\in\{2\dots T_X\}$. The accumulated cost 
is initialized with $D (1,1) = d(\zvect_{1}, \yvect_{1})$ and with $D(1,\cdot)=D(\cdot,1)=\infty$, and is then recursively estimated with:
\begin{equation}
\label{eq:-within-transition-rule}
D  (t, t')  = 
\min_{(\tau,\tau') \in\mathcal{N}_{\tau,\tau'}}
\left[ D  (t-\tau, t'-\tau') + r(\tau,\tau')d(\zvect_{t}, \yvect_{t'}) \right],
\end{equation}
where $\mathcal{N}_{\tau,\tau'}$ is the set of allowed grid transitions and $r(\tau,\tau')$ is a transition penalty.
\color{black}
$D$ is updated on the basis of minimizing over the possible between-action transitions from the \textit{past} grid points $(t-\tau,t'-\tau')$ to the \textit{current} grid point.
The best transition allowing to reach the current grid point is:
\begin{equation}
\label{eq:best-transition}
(\tau_{t}, \tau'_{t'}) = 
\argmin_{(\tau,\tau')\in\mathcal{N}_{\tau,\tau'}}  
\left[ D  (t-\tau, t'-\tau') + r(\tau,\tau') d(\zvect_{t}, \yvect_{t'}) \right].
\end{equation}
The back-pointer $\Phi$  simply stores, for the current grid point, the coordinates of the previous grid point that provided the best transition:
\begin{equation}
\Phi (t,t') = (t-\tau_{t}, t'-\tau'_{t'}).
\end{equation}
This is a general formulation that allows the implementation of a large variety of grid transitions $\mathcal{N}_{\tau,\tau'}$, e.g., \cite{RabinerJuang,ney1984use}. It is important to note that if one uses an HMM framework, this corresponds to state-to-state transitions which are dictated by the topology of the associated Markov chain. In particular, if  a Markov model is adopted, \textit{vertical} transitions are not allowed, i.e., one cannot simultaneously align $\zvect_t$ with $\yvect_{t'-1}$ \textit{and} with $\yvect_{t'}$. 
The DTW framework is more flexible than HMMs in the sense that one can implement vertical transitions, thus allowing an alignment between a long model sequence $Y$ and a short test sequence $Z$. 

In this paper we allow three types of transitions: horizontal, vertical, and diagonal, and we choose 
a weighting scheme that gives equal preference either to a diagonal transition or to a combination of horizontal and vertical transitions \cite{Mueller2007}, thus taking into account potential time discrepancies between the two sequences. The minimization is carried out over the transition pair $(\tau,\tau')$. In this paper we consider $\mathcal{N}_{\tau,\tau'}$$ = \{(0,1),(1,0),(1,1)\}$ which yields the following transition penalty to be used in (\ref{eq:-within-transition-rule}) and (\ref{eq:best-transition}):
\begin{equation}
\label{eq:r}
r(\tau,\tau') = \left\{
\begin{array}{llll}
 \tau+\tau' & \text{if}  & (\tau,\tau')\in\mathcal{N}_{\tau,\tau'}\\
 \infty & \text{else.} & &
\end{array} \right.
\end{equation}
During the backward pass of DTW, the optimal path $A^{\star}$ is found last-to-first: $(t_1,t'_1)=(T_Z,T_Y)$ and then, for $i\geq 2$ and until $(t_i,t'_i)=(1,1)$, the $i$-th assignment is simply provided by the back-pointer:
\begin{equation}
(t_i,t'_i) = \Phi(t_{i-1},t'_{i-1}).
\end{equation}
Notice that it is possible to implement this algorithm such that all the computations are done during the forward pass: The back-pointer $\Phi(t,t')$ stores the accumulated best path of grid points, rather than just the previous best grid point.

\section{Action Templates and Metaframes}
\label{section:class_templates}
In this section we describe a representation of actions that is well suited for the task of continuous recognition. The representational framework is based on the concept of templates briefly introduced in section~\ref{section:notations_definitions}. 
For each action category $l$ we define an \textit{action template} denoted by $Y^l$ and a \textit{class template} denoted by $\tilde{Y}^l$. The action template is the \textit{mean} (or the center) of the single-action examples of a class, namely the mean of $\{X_n^l\}_{n=1}^{N_l}$, and hence it is a sequence itself. A class template is a sequence of \textit{metaframes}. Each metaframe is a set of matched frames  resulting from the alignments of the action examples. 

In order to estimate the mean action of a class $l$, we seek 
the sequence $X^l_{i^\ast} \in \{X_n^l\}_{n=1}^{N_l}$ which is the closest to the class \textit{center}. This can be done in a straightforward way via the following minimization:
\begin{equation}
i^\ast = \argmin_i \sum_{j\neq i} \DTW (X_i^l, X_j^l).  \label{eq:average}
 \end{equation}
 In more detail, for each example $i$ the sum over all the alignment scores $\DTW (X_i^l,X_j^l)$ (with $j\neq i$) is computed and the example $i^{\ast}$ that minimizes this sum is selected as the class center, i.e., $X_{i^\ast}^l$, as well as the action template, i.e., $Y^l \equiv X_{i^\ast}^l$.
It is important to emphasize that this process eventually aligns each example $X_j^l$ with $Y^l$. 
\color{black}
The
class template becomes:
\begin{figure*}[t!]
\centering
\begin{tabular}{|c|c|c|c|c|c|c|}

\hline
$ \tilde{Y}^{clap}$ & $X_1^{clap}$ & $ X_2^{clap}$ & $X_3^{clap} $ &$X_4^{clap} $ &$X_5^{clap} $ & $ X_6^{clap} $ \\ 
\hline
\hline
$\tilde{y}_{1}$ &
\includegraphics[width=0.12\textwidth]{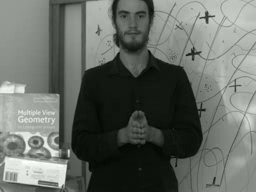} &
\includegraphics[width=0.12\textwidth]{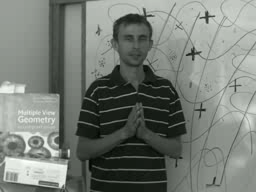} &
\includegraphics[width=0.12\textwidth]{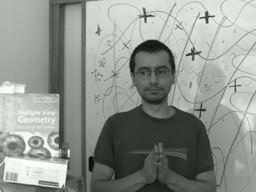} &
\includegraphics[width=0.12\textwidth]{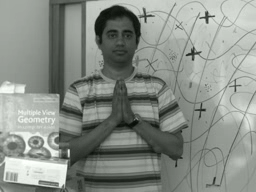} &
\includegraphics[width=0.12\textwidth]{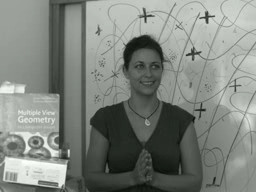} &
\includegraphics[width=0.12\textwidth]{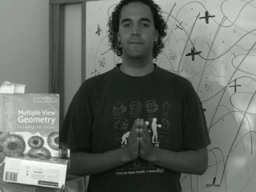} \\ 
\hline
$\tilde{y}_{5}$ &
 &
\includegraphics[width=0.12\textwidth]{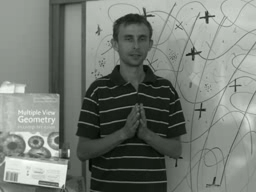} &
\includegraphics[width=0.12\textwidth]{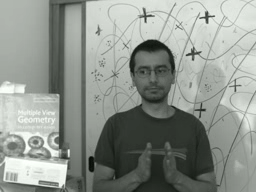} &
 &
\includegraphics[width=0.12\textwidth]{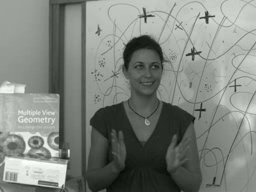} &
\includegraphics[width=0.12\textwidth]{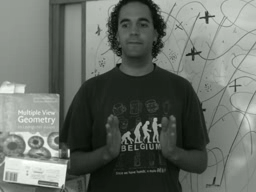} \\ 
\hline
$\tilde{y}_{10}$ &
 &
\includegraphics[width=0.12\textwidth]{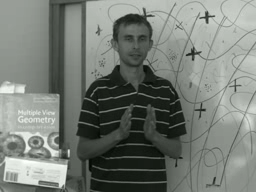} &
\includegraphics[width=0.12\textwidth]{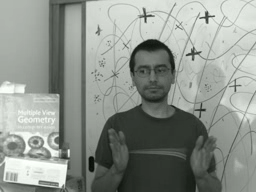} &
\includegraphics[width=0.12\textwidth]{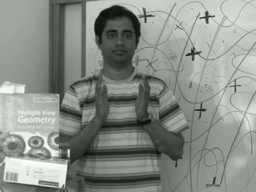} &
\includegraphics[width=0.12\textwidth]{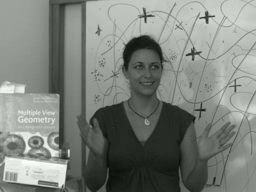} &
\includegraphics[width=0.12\textwidth]{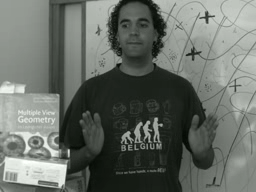}\\ 
\hline
$\tilde{y}_{15}$ &
 &
\includegraphics[width=0.12\textwidth]{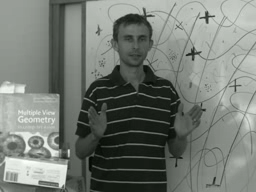} &
\includegraphics[width=0.12\textwidth]{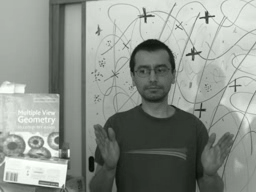} &
\includegraphics[width=0.12\textwidth]{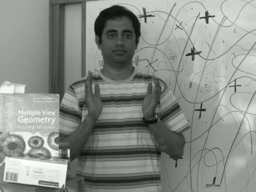} &
\includegraphics[width=0.12\textwidth]{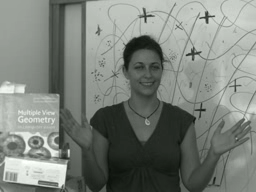} &
\includegraphics[width=0.12\textwidth]{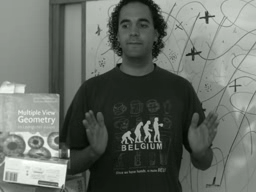}\\ 
\hline
$\tilde{y}_{20}$ &
\includegraphics[width=0.12\textwidth]{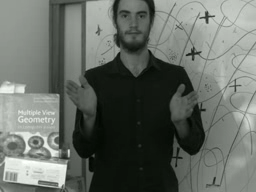} &
\includegraphics[width=0.12\textwidth]{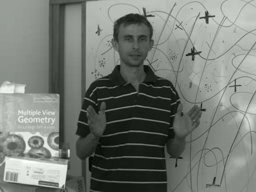} &
\includegraphics[width=0.12\textwidth]{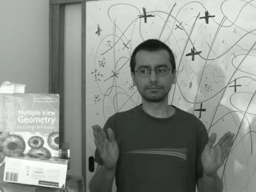} &
 &
\includegraphics[width=0.12\textwidth]{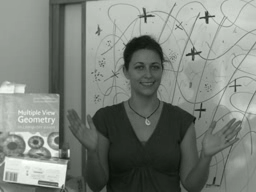} &
\includegraphics[width=0.12\textwidth]{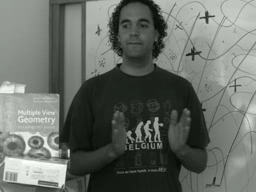}\\ 
\hline
$\tilde{y}_{25}$ &
\includegraphics[width=0.12\textwidth]{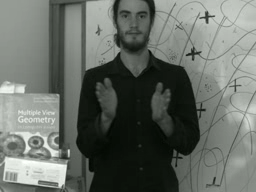} &
\includegraphics[width=0.12\textwidth]{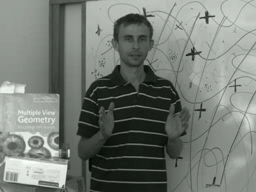} &
\includegraphics[width=0.12\textwidth]{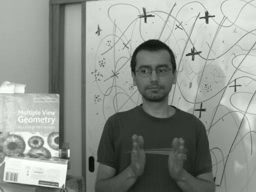} &
\includegraphics[width=0.12\textwidth]{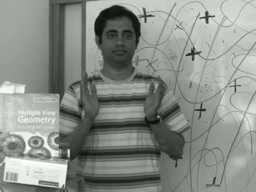} &
\includegraphics[width=0.12\textwidth]{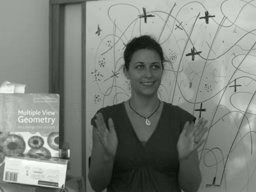} & \\ 
\hline
$\tilde{y}_{30}$ &
&
\includegraphics[width=0.12\textwidth]{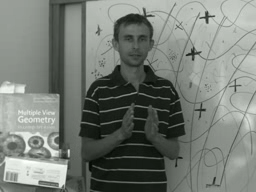} &
\includegraphics[width=0.12\textwidth]{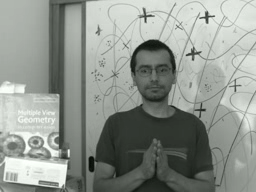} &
&
\includegraphics[width=0.12\textwidth]{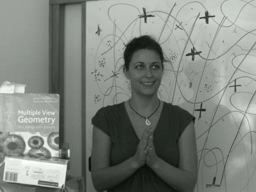} &
\includegraphics[width=0.12\textwidth]{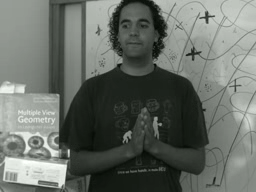}  \\ 
\hline
\end{tabular}
\caption{This figure illustrates the concepts of class-templates and of metaframes using the clap action. 
The training examples $X_1^{clap}$ to $X_6^{clap}$ are shown vertically. Each row shows the frame examples associated with a metaframe. All the examples shown in this figure are from the CONTACT dataset.
\color{black}
}
\label{fig:meta-frame}
\end{figure*}

\begin{equation}
\tilde{Y}^l = (\tilde{\yvect}_1^l, \dots \tilde{\yvect}_{t'}^l ,\dots, \tilde{\yvect}_{T_{Y^l}}^l) ,
\label{eq:model}
\end{equation}
with the same length as $Y^l$ and composed of a sequence of \textit{metaframes}, where each metaframe $\tilde{\yvect}_{t'}^l$ being a collection of matched frames resulting from the minimization (\ref{eq:average}):
\begin{equation}
\tilde{\yvect}_{t'}^l = \left\{ \xvect_{jt'}^l \right\}_{j=1}^{j=N^l_{t'}} , 
\label{eq:metaframe}
\end{equation}
where $\xvect_{jt'}^l \in X_j^l$ is associated with the ${t'}$-th metaframe, $\tilde{\yvect}_{t'}^l$, and $N^l_{t'}$ is the number of frames associated with this metaframe.
\color{black}
An example of a class template with its class center and the associated matched frames is shown on Fig.~\ref{fig:meta-frame}: there are six training examples $X_1^{clap}$ to $X_6^{clap}$ corresponding to six different actors. Once these sequences are aligned, an action template $Y^{clap}$ and a class template $\tilde{Y}^{clap}$ are obtained. Based on the frame-to-frame alignment thus obtained, each metaframe, e.g., $\tilde{y}_{1}$, $\dots$, $\tilde{y}_{30}$, is described by a varying number of example frames. 
For example, there are five frames (actors 2, 3, 4, 5, and 6) associated with metaframe $\tilde{y}_{10}$. The varying number of frames associated with a metaframe is due to ``speed of action" variabilities between different actors.
\color{black}

We now make this template-based representation well suited for continuous action recognition and we specify the notion of a
\textit{super template}, or a concatenated string of class templates, taken in an arbitrary but fixed order, namely (\ref{eq:super-template}).
The length of $\tilde{Y}^{1:L}$ is $T_{\tilde{Y}} = \sum_{l=1}^L T_{\tilde{Y}^{l}}$. Since there are length variabilities in the training set, we define a minimum and maximum temporal length associated with each class template, namely:
\begin{equation}
\label{eq:length-template}
\begin{array}{l}
\tilde{T}_{\min}^l = \displaystyle{ \min_{1\leq i \leq N^l_{t'}} } [ T_{X_i^l} ],\\
\tilde{T}_{\max}^l = \displaystyle{ \max_{1\leq i \leq N^l_{t'}} } [ T_{X_i^l} ].
\end{array}
\end{equation}

\section{Dynamic Frame Warping}
\label{section:dfw}

We consider the problem of aligning an unknown sequence $Z$ with a class template $\tilde{Y}^l$. 
Since the latter is a sequence of metaframes, the frame-to-frame distance 
$d(\xvect_t, \yvect_{t'})$ used by the dynamic time warping algorithm described in section~\ref{section:baseline_dtw} must be replaced with a \textit{frame-to-metaframe} 
distance $\tilde{d} (\zvect_t,\tilde{\yvect}^l_{t'})$. If the frames associated with a metaframe in (\ref{eq:metaframe}) obey a probability distribution function, e.g., an 
isotropic Gaussian distribution, one could easily estimate the parameters of this distribution and implement the frame-to-metaframe distance $\tilde{d}$ in closed form, i.e., a Mahalanobis distance. In 
practice there is a large variability within the frame set (\ref{eq:metaframe}) and a too simple statistical model may yield a non-discriminant metaframe description. The frame variability within a metaframe 
seems to be inherent to the bag-of-words representation which cannot guarantee that the same action performed by different people have normally distributed
descriptor vectors. Moreover, there may be temporal miss alignments introduced by the training described in section~\ref{section:class_templates} which may lead to the 
presence of mismatched frames, i.e., outliers, within a metaframe. Finally, there may not be enough training examples associated with each action such as to approximate a metaframe with a 
more sophisticated model such as a Gaussian mixture. 


For these reasons, in order to compute a frame-to-metaframe distance, we propose to adopt a reconstruction model.
Let $ \zvect_t \in \mathbbm{R}^K$ be the visual vector of a test frame and let $\tilde{\yvect}_{t'}^l$ be a metaframe composed of training examples $\{\xvect_{jt'}^l\}_{j=1}^{j=N^l_{t'}} \in \mathbbm{R}^K$. We seek to represent a test frame as a linear combination of the training frames associated with a metaframe:
\begin{equation}
 \zvect_t \approx \Xmat_{t'}^l \wvect_{t'} ,
\label{eq:frame-to-metaframe}
\end{equation}
where $\Xmat_{t'}^l = [ \xvect_{1t'}^l \hdots \xvect_{jt'}^l \hdots \xvect_{N^l_{t'}t'}^l]$ is a $K\times N^l_{t'}$ matrix and $\wvect_{t'}=(w_{1t'} \hdots w_{N^l_{t'}t'} )\tp$ is the vector of reconstruction coefficients. 
In order to avoid over-smoothing and because only a few training frames are likely to be similar to the test frame,
we seek a sparse solution for $\wvect_{t'}$ by solving the following \textit{basis pursuit denoising} problem \cite{Chen2001}:
\begin{equation}
\wbarre_{t'}=\argmin_{\wvect} \|\zvect_t -  \Xmat_{t'}^l \wvect  \|_2+\gamma\|\wvect\|_1 .
\label{eq:basis_pursuit}
\end{equation}
The $\ell_1-$norm regularizer ensures the sparseness of the solution while the value of $\gamma$ can be tuned such that a satisfactory tradeoff is achieved between the sparsity level and the value of the Euclidean norm. The above problem casts into a convex quadratic programming method which can be efficiently solved, e.g., \cite{BoydBook}.
Notice that the minimization (\ref{eq:basis_pursuit}) yields a solution even if the linear system (\ref{eq:frame-to-metaframe}) is under determined, namely if $N^l_{t'}<K$.


The solution yielded by (\ref{eq:basis_pursuit}) is a sparse coefficient vector 
with the set of non-zero indices being defined by $\mathcal{S}=\{j | j\in\{1,\dots\,N^l_{t'}\}, \overline{w}_{jt'} \neq 0\}$. As is customary in basis pursuit~\cite{BoydBook}, the sparsity is subsequently reused to compute a normalized cost, since the minimum cost of (\ref{eq:basis_pursuit}) can be biased favoring too sparse solutions. Therefore, we also build a reduced $K\times |\mathcal{S}|$ matrix whose columns are the remaining visual vectors $\Xbarre_{t'}^l=[\xvect^l_{jt'}], {j\in \mathcal{S}}$ 
and we solve the following minimization problem that yields a frame-to-metaframe distance in closed-form \cite{Evangelidis08}:
\begin{equation}
\label{eq:frame-to-metaframe-dis}
\tilde{d} (\zvect_t,\tilde{\yvect}^l_{t'}) = \min_{\wvect} \left\| \zvect_t - \frac{\Xbarre_{t'}^l \wvect}{\| \Xbarre_{t'}^l \wvect \|_2} \right\|_2^2 \; s.t. \sum_{i=1}^{|\mathcal{S}|} {w}_{i}=1.
\end{equation}
The alignment between a test sequence $Z$ and a class template $\tilde{Y}^l$, namely $A^{\ast}(Z,\tilde{Y}^l)$, is estimated by replacing $d$ with $\tilde{d}$ in (\ref{eq:dtw}); the corresponding dissimilarity associated with an optimal path becomes:
\begin{equation}
\DFW (Z,\tilde{Y}^l) = \frac{1}{|A^\ast|} \sum_{i=1}^{|A^\ast|} \tilde{d} (\zvect_t,\tilde{\yvect}^l_{t'}).
\label{eq:dfw_distance}
\end{equation}
Sequence alignment based on frame-to-metaframe distances will be referred to as \textit{dynamic frame warping} (DFW).

The above formulation allows a straightforward implementation of \textit{isolated action recognition} in the spirit of nearest-neighbor search \cite{Ikizler:2009,BrendelT10} but in a slightly more efficient way. Indeed, if a test sequence $Z$ contains a single unknown action, the class label can be recovered by:
\begin{equation}
\label{eq:isolated-recognition}
l^\ast = \argmin_{l\in\mathcal{L}}  \DFW(Z,\tilde{Y}^l) .
\end{equation}
\color{black}

\section{Continuous Recognition}
\label{section:continuous_recognition}
The dynamic programming framework and template-based representation just described can be used to simultaneously segment an unknown sequence into isolated actions and to recognize them. More precisely, the test sequence $Z$ is composed of an unknown number of, possibly repeating, actions in an unknown order. Moreover, not only that the between-action boundaries are not known in advance, but the transitions from one action to the next one are often smooth; this further complicates the task of segmentation.
In this section we describe two extensions of the baseline dynamic frame warping (DFW) method outlined in section~\ref{section:dfw}, namely the one-pass DFW and two-pass DFW algorithms. 

Both proposed methods constitute elegant extensions of dynamic time warping for isolated recognition. Isolated recognition only requires \textit{within-action transitions} between frames, e.g., (\ref{eq:-within-transition-rule}) and associated rules (\ref{eq:r}). In continuous recognition \textit{between-action transitions} are required as well. The first algorithm, OP-DFW, implements a mechanism allowing either to stay in the same template or to jump from the end-frame of a template to the begin-frame of another template. 
This is somehow equivalent to an HMM and to the Viterbi algorithm that allows jumps between HMM states. The main advantage of the proposed DTW-like algorithm is that there is no need to compute partition functions as with probabilistic graphical models.
\color{black}

The second algorithm, TP-DFW, implements continuous recognition quite differently: an action recognition stage (or the first pass) is followed by a sequence segmentation stage (or the second pass). Firstly, isolated recognition (\ref{eq:isolated-recognition}) is applied to a sub-sequence $Z_{t_b:t_e}$ of $Z$ parameterized by a begin-frame $t_b$ and an end-frame $t_e$. Isolated recognition is repeatedly applied to each possible sub-sequence such that best labels and alignment scores are estimated for all possible sub-se\-quen\-ces. Secondly, a dynamic programming procedure is applied to the $(t_b,t_e)$ grid that was generated during the first pass, in order to find an alignment path that coincides with an optimal string of labels based on the previously computed scores. 

Prominent features of both methods are that they can deal with test sequences having an arbitrary number of actions and that they allow a large variability in terms of action durations. 
The first pass of the TP-DFW method can be carried out by virtually \textit{any} isolated recognition method and hence it is suitable for combining continuous recognition with a discriminative method.
\color{black}


\subsection{One-Pass Dynamic Frame Warping}
\label{sec:op-dfw}
\begin{figure*}[t!]
\centering
\begin{tabular}{cc}
\includegraphics[width=0.4\textwidth]{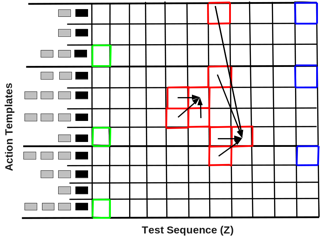} &
\includegraphics[width=0.4\textwidth]{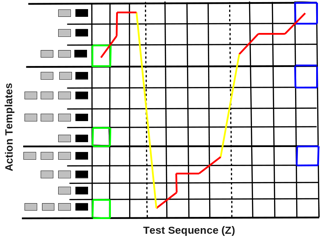}\\
{(a) Within-action and between-action transitions}&  
{(b) The alignment path extracted by backtracking} \\
\end{tabular}
\caption{\label{fig:1stageDP} This figure illustrates the forward (a) and the backward steps (b) of the OP-DFW algorithm. Each square denotes a grid point $(t,t'(l))$. (a) The forward step is initialized at the first frame $(1,t'(l))$ of each action template, shown by green squares. The accumulated cost $\tilde{D}$, the bakpointers $\tilde{\phi}$ and the length constraint $\tilde{T}$ are updated at each grid point by enforcing the \textit{between-action} and the \textit{within-action} transitions as indicated by the red squares. (b) The alignment path, where alignments within actions are shown in red and the action jumps are shown in yellow. The path is initialized at one of the blue squares, i.e, one of the last frames of the action templates, and is backtracked. The dotted vertical grid lines materialize the estimated action boundaries.}
\end{figure*}

We consider the task of finding an optimal path between un unknown sequence $Z$ and a super-template $\tilde{Y}^{1:L}$, defined in (\ref{eq:super-template}), and as explained in section~\ref{section:class_templates}. 
The optimal path will contain the information needed to split the sequence $Z$ into sub-sequences and to assign an action label to each sub-sequence frame, as defined by (\ref{eq:Z-segmentation}).
\color{black}
We must modify the alignment definition (\ref{eq:assignment-def}) to allow for 
 frame-to-\textit{metaframe} assignments:
\begin{equation}
\label{eq:assinment-metaframe}
\tilde{A}(Z,\tilde{Y}) = \left\{ (t_i, t'_i (l)) \right\}_{i=1}^{i=|\tilde{A}|}
\end{equation}
where $t_i$ is the frame index of $Z$, $1\leq t_i \leq T_Z$, $l$ is the template index, $1\leq l \leq L$, $t'_i(l)$ is the frame index of template $Y^l$, $1\leq t' _i(l)\leq T_{Y^l}$, and $i$ is the path index. The optimal path is the solution of the following optimization problem:
\begin{equation}
\label{eq:continuous-dtw}
\left\{
\begin{array}{l}
\tilde{A}^{\ast}(Z,\tilde{Y}) = \argmin_{\tilde{A}} \sum_{i=1}^{|\tilde{A}|} \tilde{d} (\zvect_{t_i},\tilde{Y}^{l_i}_{t'_i})\\
\begin{array}{ll}
\text{s.t. } &  \left\{ \begin{array}{l} (t_1,t_1') =  (1,1) \\ 
 (t_{|\tilde{A}|},t_{|\tilde{A}|}') = (T_Z,T_{\tilde{Y}}), \end{array} \right.
 \end{array}
 \end{array}
 \right.
\end{equation}
\color{black}
where the frame-to-metaframe distance $\tilde{d}$ was defined in (\ref{eq:frame-to-metaframe-dis}). There is a similar expression for the DFW dissimilarity statistics associated with the optimal path:
\begin{equation}
\DFW (Z,\tilde{Y}) = \frac{1}{|\tilde{A}^\ast|} \sum_{i=1}^{|\tilde{A}^\ast|}  \tilde{d} (\zvect_{t_{i}^{\ast}},\tilde{Y}^{l_{i}^{\ast}}_{t'_{i}{^{\ast}}}).
\end{equation}
The \textit{within-action} transition rule (\ref{eq:-within-transition-rule}) defined in the case of isolated recognition must be augmented with a \textit{bet\-ween-action} transition (or jump) rule. Let $\tilde{D}(t,t'(l))$ be the accumulated cost associated with the grid point $(t,t'(l))$. As before, we also define a back-pointer $\tilde{\Phi}(t,t'(l))$ at each grid point. In addition we define an accumulated temporal length $\tilde{T}(t,t'(l))$ that is associated with each grid point and with each template.

During the forward step, once initialized, the accumulated costs $\tilde{D}$, back-pointers $\tilde{\Phi}$ and lengths $\tilde{T}$ are estimated at each grid point $(t,t'(l))$ while enforcing two types of transitions. The algorithm can be outlined as follows:
\begin{itemize}
\item \textbf{Initialization.} The unknown sequence starts with the first frame of any of the class templates, hence, the cost is initialized with:
\begin{equation}
\tilde{D}(1, t'(l)) = \left\{
\begin{array}{ll}
\tilde{d} (\zvect_1,\tilde{Y}^l_{1}) & \text{if } t'(l)=1\\
\infty  & \text{if } 2\leq t'(l) \leq T_{\tilde{Y}^l}
\end{array} \right.
\end{equation}
while the temporal length is equal to $1$ at the start of each class template, $\tilde{T}(1,1(l))=1$.
\item \textbf{Within-action} transitions occur between two grid points belonging to the same template $l$, i.e., at $2\leq t \leq T_{Z}, 2\leq t'(l) \leq T_{Y^l}$, $ 1\leq l \leq L$, and they are strictly identical to the isolated case:
\begin{equation}
\label{eq:within-frame-rule}
\tilde{D}(t,t'(l)) = \min_{\tau,\tau'} [ \tilde{D}(t-\tau,t'(l)-\tau') + r(\tau,\tau')\tilde{d} (\zvect_t,\tilde{Y}^l_{t'})]
\end{equation}
where the transition penalty $r$ was defined in (\ref{eq:r}). The optimal transition is given by:
\begin{equation}
(\tau_{t}, \tau'_{t'}) = \argmin_{\tau,\tau'}  [ \tilde{D}  (t-\tau, t'(l)-\tau') + r(\tau,\tau') \tilde{d}(\zvect_{t}, \tilde{Y}^l_{t'}) ].
\end{equation}
The back-pointer indicates that the path must stay within the same action $l$:
\begin{equation}
\tilde{\Phi} (t,t'(l))= (t-\tau_{t}, t'(l) - \tau'_{t'}).
\end{equation}
The accumulated temporal length is recursively updated using the following rule:
\begin{equation}
\tilde{T}(t,t'(l)) = 
\tilde{T}(t-\tau, t'(l) -\tau') + \tau_{t}. 
\end{equation}
Therefore, the accumulated temporal length of a template does not take into account vertical transitions, i.e., $\tau_{t}=0$, and is incremented only if either a horizontal or a diagonal within-action transition is selected $\tau_{t}=1$. 
\item \textbf{Between-action} transitions can only happen between the end-frame of any template $k, 1\leq k\leq L$ and the begin-frame of the current template $l$, i.e., from a grid point $(t-1, T_{\tilde{Y}^k})$ to the current grid point $(t,t')=(t,1(l))$, $\forall k, 1\leq k \leq L$ (thus allowing transitions from the end-frame of a template $l$ the begin-frame of the same template $l$, i.e., the same action is repeated). 

First, the best template label $k^{\ast}$ is estimated: \begin{equation}
k^\ast = \displaystyle {\argmin_{1\leq k \leq L} }[\tilde{D}(t-1,T_{\tilde{Y}^k}) + r_k  \tilde{d} (\zvect_t,\tilde{Y}^l_{1}) ] 
\end{equation}
and the associated between-action transition is then selected:
\begin{equation}
\tilde{D}(t,1(k^{\ast})) = \tilde{D}(t-1,T_{\tilde{Y}^{k^\ast}}) + r_{k^\ast}  \tilde{d} (\zvect_t,\tilde{Y}^l_{1}) 
\end{equation}
where the transition penalty $r_{k}$ is defined such that a sub-sequence of $Z$ having an accumulated temporal length that is too short or too long is disregarded:
\begin{equation}
r_k = \left\{
\begin{array}{ll}
1 & \text{if } \tilde{T}_{\min}^k \leq \tilde{T}(t,T_{\tilde{Y}^k}) \leq \tilde{T}_{\max}^k \\
\infty & \text{otherwise}.
\end{array} \right.
\end{equation}
Second, the accumulated between-action cost is estimated with:
\begin{equation}
\label{eq:begin-frame-rule}
\tilde{D}(t,1(l))  =  \min \big[ \tilde{D}(t-1,1(l)) + \tilde{d} (\zvect_t,\tilde{Y}^l_{1}), 
\tilde{D}(t,1(k^{\ast}))\big].
\end{equation}

The associated $\argmin$ function returns a template label $\delta$ that is equal either to $k^{\ast}$ if a between-action transition from the end-frame of $k^{\ast}$ to the start-frame of $l$ is preferred by (\ref{eq:begin-frame-rule}) or to $l$ otherwise. 
Finally, in this case the back-pointer must indicate whether a between-action transition occurs or not:
\begin{equation}
\tilde{\Phi} (t,1(l))= (t-1, t'(\delta))
\end{equation}
with $t'(\delta)=T_{\tilde{Y}^{k^{\ast}}}$ if $\delta=k^{\ast}$ (there is a between-action transition) or $t'(\delta)=1(l)$ if $\delta = l$ (there is no between-action transition).
\end{itemize}
During the backward step of the OP-DWF, the optimal path $\tilde{A}^\ast$ is found from-last-to-first. The path is initialized with:
\begin{equation}
(t_1, t'_1( l)) = \argmin_{1\leq l \leq L} \tilde{D} (T_Z, T_{\tilde{Y}^l})
\end{equation}
and then for $i\geq 2$ and until $(t_i, t'_i(l)) = (1,1(l)), \forall l$:
\begin{equation}
(t_i, t'_i(l)) = \tilde{\Phi} (t_{i-1}, t'_{i-1}(l)).
\end{equation}
Once the optimal path is thus determined, it is straightforward to obtain a partitioning of the unknown sequence $Z$ into $J$ sub-sequences, i.e., (\ref{eq:Z-segmentation}) such that each sub-sequence is optimally aligned with a class template. The OP-DFW algorithm is sketched on Figure~\ref{fig:1stageDP}.

\subsection{Periodic Actions}
\label{sec:periodic-actions}
\addnote[periodic]{1}{Many human actions contain some kind of periodicity, e.g., running, walking, waving, etc., and the continuous (classification and segmentation) recognition of this type of actions must be carefully addressed. Periodic actions are generally composed of \textit{motion patterns} that are repeated an arbitrary number of times in the same action. Both training and recognition of these actions must be carefully addressed. The proposed OP-DFW algorithm handles periodic actions in the following way.
Firstly, we need to learn a class template for each motion-pattern, i.e., section \ref{section:class_templates}. This is done by carefully annotating periodic-action examples in order to obtain a training set for each motion pattern and learn a \textit{motion-pattern template}. Secondly, OP-DFW is applied to an unknown test sequence. Whenever, a periodic action is present in the test sequence, the algorithm jumps from one motion-pattern to the next motion-pattern (which corresponds to a between-action transition) or stays within the same motion-pattern (which correspond to a within-action transition). Fig.~\ref{fig:continuous_rec} illustrates the behavior of the OP-DFW method with two different annotations. The top-right plot shows the alignment path where each class-template is associated with a periodic action, or \textit{per action annotation}. In this case, periodic actions are modeled like any other action and the test sequence is segmented into three actions, namely ``clap-wave-punch". The top-left plot shows the alignment path where each class-template is replaced with a motion-pattern template, or \textit{per motion-pattern annotation}. In this case the test sequence is segmented into nine motion patterns, namely ``clap-clap-clap", ``wave-wave-wave", and ``punch-punch-punch". Note that if the test sequence is composed of a single periodic action (this corresponds to isolated action recognition), the OP-DFW treats it as a sequence of patterns and therefore achieves classification and segmentation. Therefore, OP-DFW can deal with periodic actions composed of an arbitrary number of repeated patterns of different types.
}

\subsection{Two-Pass Dynamic Frame Warping}
\label{subsection:TPDFW}
\label{ssec:TP-DFW}
\begin{figure*}[t!]
\centering
\begin{tabular}{cc}
\includegraphics[width=0.4\textwidth]{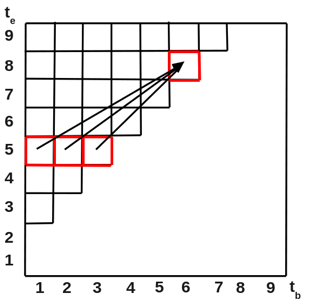} &
\includegraphics[width=0.4\textwidth]{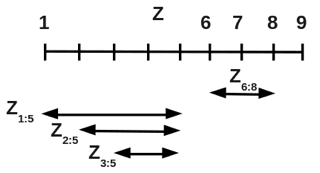}\\
(a) Between-action transitions in TP-DFW &  
(b) Sub-sequence representation \\
\end{tabular}
\caption{This figure illustrates the \textit{sequence-level pass} of the TP-DFW algorithm with a test sequence $Z_{1:9}$ and $T_{\min}=2$. The \textit{action-level pass} (which is not shown) assigns an action label (\ref{eq:best-labels}) and an alignment cost~(\ref{eq:best-dtw}) at each grid point, i.e., at each possible sub-sequence of $Z_{1:9}$. (a) An example of three possible transitions from the sub-sequences ending in $t_e=5$ to a sub-sequence starting in $t_b=6$, namely from $Z_{1:5}$, $Z_{2:5}$, or $Z_{3:5}$ to $Z_{6:8}$. (b) A different way of representing the possible transitions between actions ending at $t_e=5$ and actions starting at $t_b=6$.}
\label{fig:2stageDP}
\end{figure*} 

We consider now a sub-sequence $Z_{t_b:t_e}$ of the test sequence $Z_{1:T_Z}$ parameterized by its first and last frames. 
The ordered sets of begin-frame indexes $\{t_b\}_{1\leq t_b < T_Z}$ and of end-frame indexes $\{t_e\}_{1< t_e \leq T_Z}$ are such that 
$1\leq t_b < t_e\leq T_Z $. 
We impose the additional constraint that $t_e - t_b \geq T_{\min}$, i.e., $T_{\min}+1$ is the minimum number of frames in an action. TP-DFW proceeds in two passes, as follows.
\begin{itemize}
\item \textbf{The action-level pass} performs repeated estimations of (\ref{eq:isolated-recognition}) such as to obtain an alignment cost for each class $l$ and for each sub-sequence $Z_{t_b:t_e}$:
\begin{equation}
\label{eq:beginend-label}
\mathcal{C}(l,t_b,t_e) = \DFW(Z_{t_b:t_e},\tilde{Y}^l)
\end{equation}
from which we can obtain the best class label $\hat{l}$ for each sub-sequence with the associated cost:
\begin{eqnarray}
\label{eq:best-labels}
\hat{l} (t_b,t_e) &=& \argmin_{l\in\mathcal{L}} [ \mathcal{C} (l,t_b,t_e)  ], \\
\label{eq:best-dtw}
\hat{d} (t_b,t_e) &=& \min_{l\in\mathcal{L}} [ \mathcal{C}(l,t_b,t_e) ].
\end{eqnarray}
Equation (\ref{eq:beginend-label}) implies that an optimal path is sought for each sub-sequence/template pair $(Z_{t_b:t_e}, \tilde{Y}^l)$ and hence the dynamic frame warping algorithm outlined in section~\ref{section:dfw} must be 
applied to
\color{black}
each $(l,t_b,t_e)$ triplet. Let $(t,t'(l))$ be the current grid point associated with each sub-sequence/template alignment, hence $t_b \leq t \leq t_e$ and $1 \leq t'(l) \leq T_{\tilde{Y}^l}$. The corresponding accumulated cost $D$ is initialized as already explained in section~\ref{section:baseline_dtw}, namely $D(t_b,1(l)) = \tilde{d}(z_{t_b},\tilde{y}_1^l)$ for each action class $l\in \mathcal{L}$, while the remaining grid points of the first row and first column are initialized to $\infty$. The forward step remains strictly identical to the one explained in detail in section~\ref{section:baseline_dtw}. The backward step of this dynamic programming process builds a path from-last-to-first starting at $(t_e,T_{\tilde{Y}^l})$ and necessarily ending at $(t_b,1(l))$. This allows to estimate a table of labels $\hat{l}$ and a table of alignment costs $\hat{d}$ indexed by $(t_b,t_e)$, i.e., (\ref{eq:best-labels}) and (\ref{eq:best-dtw}).

\item \textbf{The sequence-level pass} takes as input these two tables, both indexed by $(t_b,t_e)$, and attempts to align the unknown sequence with a string of action templates such as to minimize an overall accumulated distance and such that an optimal partitioning of the form of (\ref{eq:Z-segmentation}) is eventually obtained. For this purpose, dynamic programming is invoked as follows. Let $t_b$ and $t_e$ \textit{be the horizontal and vertical indexes of a DP grid}, e.g., Fig.~\ref{fig:2stageDP}. Hence, there is a sub-sequence $Z_{t_b:t_e}$, a cost $\hat{d}$ and a label $\hat{l}$
associated with each
grid point $(t_b,t_e)$, with $1\leq t_b < t_e \leq T_Z$. The task of the sequence-level DP pass is to find an optimal path, namely:
\begin{equation}
\hat{A}^\ast = \argmin_{\hat{A}} \sum_{j=1}^{|\hat{A}|} \hat{d} (t_{b,j},t_{e,j}).
\label{eq:two-level-optimal}
\end{equation}
The accumulated cost of this DP process is estimated with:
\begin{eqnarray*}
&&\hat{D}(t_b,t_e) =  \\
&& \left\{
\begin{array}{ll}
\hat {d} (t_b,t_e) + \displaystyle{ \min_{k \in [1, t_b^{\min}]}} \hat{D}(k,t_b - 1),
& \text{ if } t_e \geq t_b^{\max}  \\
\infty,  & \text{ otherwise}.
\end{array}
\right.
\end{eqnarray*}
with $t_b^{\min} = t_b - T_{\min}$ and $t_b^{\max} = t_b + T_{\min}$. The forward step applies the above transition rule to each grid point. The backward step finds an optimal path $\hat{A}^{\ast}=\{t_{b,j},t_{e,j}\}_{j=1}^{J}$ with $J=|\hat{A}^{\ast}|$. Since there is an action label associated with each grid point, the path $\hat{A}^{\ast}$ corresponds to an optimal segmentation of the unknown sequence $Z$ as well as to a synthesized string of $J$ class templates:
\begin{equation}
\left\{
\begin{array}{l}
Z_{1:T_Z} = (Z_{t_{b,1}:t_{e,1}}, \hdots Z_{t_{b,j}:t_{e,j}}, \hdots Z_{t_{b,J}:t_{e,J}}) \\
\\
\tilde{Y}^{l_1:l_J} =  (\tilde{Y}^{l_1}, \hdots \tilde{Y}^{l_j}, \hdots \tilde{Y}^{l_J}).
\end{array}
\right.
\end{equation}
\end{itemize}
The sequence-level pass of TP-DFW is illustrated on Fig.~\ref{fig:2stageDP}.

\subsection{Method Complexity}
\label{sec:complexity}
We now analyze in detail the algorithmic complexity of the proposed methods. We start by analyzing the complexity of isolated recognition using the dynamic frame warping (DFW) algorithm and then we discuss continuous recognition based on OP-DFW or on TP-DFW algorithms. Without loss of generality, we only consider the runtime procedures. Let a super-template $\tilde{Y}^{1:L}$ be composed of $L$ class-templates $\tilde{Y}^l,1\leq l \leq L$ and let $T_Y$ be the average length of a class-template. Moreover, let $N$ be the average number of frame examples associated with a metaframe. Hence, each class-template contains, on an average, $T_YN$ frames, while a super template contains $LT_YN$ frames. Moreover, let $Z$ be a test sequence with $T_Z$ frames. As before, we denote with $K$ the dimension of the features vectors associated with the per-frame descriptors.

For each action, the forward step of DFW computes a frame-to-metaframe distance and an accumulated cost on a $T_Y \times T_Z$ grid. The frame-to-metaframe distance is computed using (\ref{eq:basis_pursuit}) and (\ref{eq:frame-to-metaframe-dis}). The former equation needs an iterative solver while the latter can be solved in closed form. Efficient solutions for (\ref{eq:basis_pursuit}) were suggested, such as the exact solver~\cite{Gill2011} or the greedy solver~\cite{Trop2007} (orthogonal matching pursuit, or OMP). Notice that OMP suits well in our case and its complexity is $O(|\mathcal{S}| KN)$. Once the sparse solution is available, the computation of a frame-to-metaframe distance in (\ref{eq:frame-to-metaframe-dis}) implies the solution of a $|\mathcal{S}|\times|\mathcal{S}|$ linear system~\cite{Evangelidis08} whose complexity is $O(|\mathcal{S}|^3)$. Since (\ref{eq:frame-to-metaframe-dis}) yields a sparse solution, $|\mathcal{S}| \ll K,N$ and therefore we have $|\mathcal{S}| KN \ll |\mathcal{S}|^3$. To conclude, in the presence of $L$ action categories, the complexity of isolated recognition based on DFW approximately is 
\begin{equation}
\label{eq:DFW-complex}
O(|\mathcal{S}| KN L T_Y T_Z + L T_Y T_Z).
\end{equation}
The only difference between DFW and OP-DFW is that it
incorporates the computation of between-action transitions. Between-action transitions are computed at grid points corresponding to the last frame of a class template. Since there are $L$ class-templates, the between-action transitions are estimated $LT_Z$ times. Hence the complexity of the forward step of OP-DFW approximately is:
\begin{equation}
\label{eq:OP-complex}
O(|\mathcal{S}| KN L T_Y T_Z + L T_Y T_Z + L T_Z).
\end{equation}
The TP-DFW algorithm is more complex for two reasons. First it necessitates two dynamic programming passes: an action-level pass and a sequence-level one. Nevertheless, it can be easily seen from the description of the method that the first pass necessitates more computations than the second one. Let $M$ be the number of sub-sequences of the test sequence $Z$ and let $T_S$ be the average length of a sub-sequence. Since $t_b < t_e$, there are $T^2_Z/2$ sub-sequences in $Z$. However, because we only consider sub-sequences of a minimum length, $T_{\min}$, the number of sub-sequences is $(T_Z-T_{\min})(T_Z-T_{\min} +1)/2$. The method starts by computing a grid of frame-to-metaframe distances for all the class-templates, followed by an accumulated-cost grid for each sub-sequence and for each class-template. Hence the complexity of the action-level pass of TP-DFW approximately is $O(L|\mathcal{S}| KN T_Y T_Z + ML T_Y T_S)$. Moreover, the sequence-level pass computes a cost at each grid point $(t_b,t_e)$, with $1\leq t_b < t_e \leq T_Z$. Therefore the complexity of TP-DFW approximately is:
\begin{equation}
\label{eq:TP-complex}
O(|\mathcal{S}| KN L T_Y T_Z + ML T_Y T_S + T_Z^2/2).
\end{equation}
By inspection of (\ref{eq:DFW-complex}) and (\ref{eq:OP-complex}) one can immediately observe that the proposed OP-DFW method is barely more complex that performing isolated recognition within the same framework. The complexity of TP-DFW is higher because the algorithm considers a large number of sub-sequences of the test sequence. Notice however that the second (action-level) pass does not introduce a substantial computation burden.
\color{black}
\subsection{Null-Class Representation}
\label{sec:null-class}
\addnote[null-class]{1}{Robust recognition should be able to deal with videos that include actions that are not labeled in the training dataset, or with truncated actions. This can be done by introducing a \textit{null class}. We adopt a null-class model inspired from speech and that can be easily incorporated in our methodology. Likewise continuous action re\-cognition, continuous speech entails the recognition of sequences of spoken words corrupted by the presence of noise, background sounds, prosody (non-speech sounds emitted by the spea\-ker), out-of-vocabulary words, badly pronounced words, words from another language, truncated words, etc. It is common practice to model environmental and non-speech sounds with a single-state HMM and out-of-vocabulary words with multi-state HMMs thus modeling their constituent sub-units, e.g., phonemes \cite{gales2008application}. Sub-unit modeling could also be beneficial for recognizing truncated actions. 
We already discussed in Section~\ref{section:introduction} that it is neither practical nor easy to model actions using sub-units, with the notable exception of periodic actions, i.e., section~\ref{sec:periodic-actions}.}

Therefore we propose to model the null class as a template of unit length, i.e., composed of a single metaframe. All the frames in the training data which do not belong to any of the \textit{in-vocabulary} actions are assigned to the null-class metaframe. The representation of a test frame in terms of a sparse set of model frames, i.e., section~\ref{section:dfw}, allows to robustly estimate a test-frame-to-null-class distance and hence to account for the large variability of the null-class examples.
Therefore, there is no need to consider within-action transitions inside the null-class. Between-action transitions allow to jump back and forth between the null-class and any action class; they also enable the null-class to successively jump onto itself, thus allowing sequences of arbitrary length to be assigned to the null-class. 
\color{black}
\section{Experimental Results}
\label{section:experiments}
\label{sec:experiments}
In this section we present in detail results obtained with our method which we compare with two recently published continuous recognition methods \cite{HoaiLD11,ShiWanCheSmoijcv11}. The reported results were obtained using the Matlab implementations provided by the authors of these two methods. 

We also compare our method with a state-of-the-art  HMM implementation \cite{young2009htk} plugged into both the one-pass and two-pass methods.
As outlined before, the proposed algorithms perform se\-quence-level segmentation and action-level classification, therefore they provide a label for each frame in the test sequence. Accordingly, the performance of the algorithms is quantified through the percentage of true positives (correctly labeled frames). 
\color{black}

\subsection{Datasets}
We report experiments with 
three publicly available datasets, Hollywood-1 \cite{LaptevMSR08},  Hollywood-2 \cite{marszalek2009actions}, 
\color{black}
RAVEL\footnote{\url{http://perception.inrialpes.fr/datasets/Ravel/}} \cite{ASWFCKDH13}, as well as with a dataset specifically gathered and annotated for evaluating continuous action recognition methods: the \textit{continuous action} (CONTACT) dataset.

The CONTACT dataset consists of three periodic actions, i.e. \textit{clap}, \textit{punch} and \textit{wave}, that are continuously performed by seven actors in a predetermined order, clap-punch-wave, e.g., Fig.~\ref{fig:continuous_rec}. 
Since these are periodic actions formed of repetitive motion patterns, there are two possible ways of annotating the boundaries: between-action annotation and between-motion annotation. Both these boundaries are similarly handled by our one-pass method (see section~\ref{sec:periodic-actions}).
Fig.~\ref{fig:continuous_rec} shows that the results obtained with these two annotations are similar. 
\color{black}

\begin{figure}[t!]
\begin{center}
\begin{tabular}{cc}
\includegraphics[width=0.48\columnwidth]{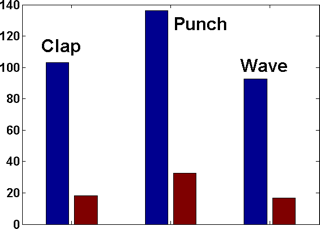} &
\includegraphics[width=0.48\columnwidth]{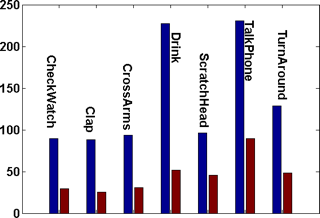} \\
{(a)CONTACT dataset}&
{(b) Ravel dataset}\\
\includegraphics[width=0.48\columnwidth]{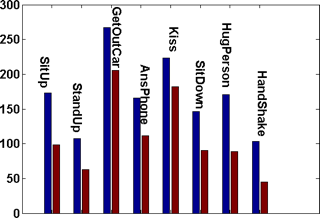}&
\includegraphics[width=0.46\columnwidth]{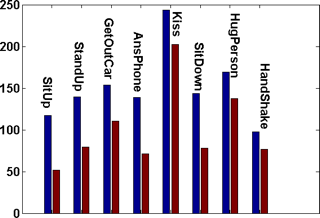}\\
{(c)Hollywood-1 training set}&
{(d) Hollywood-1 test set}\\ & \\
\includegraphics[width=0.40\columnwidth]{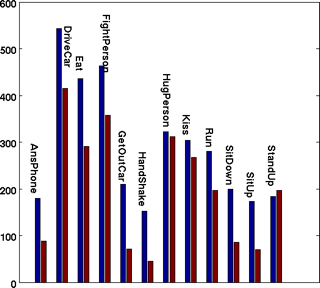}&
\includegraphics[width=0.40\columnwidth]{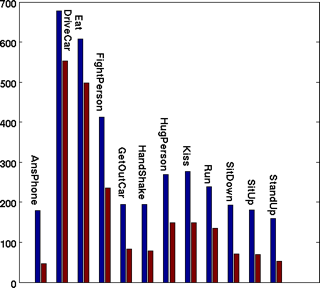}\\
{(e)Hollywood-2 training set}&
{(f) Hollywood-2 test set}\\
\end{tabular}
\end{center}
\caption{The mean (blue) and the standard deviation (red) of the temporal lengths for each dataset and for each action category. }
\label{fig:AvgVarDataSets}
\end{figure}

The RAVEL dataset has been collected for training human-robot interaction (HRI) tasks~ \cite{ASWFCKDH13}. The actors were prompted to repeat seven actions: Check-watch, Cross-arms, Scratch-head, Phone-talk, Turn-around, and Drink in a random order. Each set of actions was repeated three times by twelve actors. The actors were not given any particular instructions on how to perform the actions other than coming to a rest position after each action. Consequently, this leads to large intra-action variabilities and to smooth inter-action transitions. This is in strong contrast with the abrupt inter-action transitions associated with the artificially concatenated actions obtained from the isolated-action datasets, e.g., Hollywood-1.

The Hollywood-1 dataset consists in video samples extracted from 32 movies, split in 219 and 211 videos for training and testing respectively. Eight actions are performed in total: Sit-down, Sit-up, Ans-phone, Hug-person, Kiss, Get-out-car, Hand-shake and Stand-up. Since each sample (video) contains a single action, we artificially concatenated isolated-action videos, thus generating $30$ test sequences, each being composed of $8$ randomly selected actions in an arbitrary order. We trained our method with a subset of actions, namely Answer-phone, Hug-person, Kiss, and Sit-down. The rest of the actions were treated as belonging to a null (see section~\ref{sec:null-class} and \cite{HoaiLD11}).

\addnote[hollywood-2]{1}{The Hollywood-2 dataset \cite{marszalek2009actions} consists in a collection of videos corresponding to twelve isolated actions. These videos were manually extracted from 69 different movies. There are 823 manually annotated training examples (clean training set), 810 manually annotated test examples (clean test set), as well as 810 automatically annotated training examples (script training set). For training, we only used the clean training set  while for testing we used both the clean and script training sets. Generally speaking, Hollywood-2 is a very challenging dataset because the action boundaries are not very accurately annotated in both the clean and training sets. This will affect both the continuous and isolated recognition results, because our sequence alignment method remains affected by the presence of truncated actions in the test data.}

Fig.~\ref{fig:AvgVarDataSets} summarizes the mean temporal lengths of the action categories and their standard deviation for all these datasets. As it can be seen, there are large length variations associated with the three dataest.

\subsection{Implementation}
\label{ssec:implementation}
State-of-the-art methods for frame coding suggest the use of bag-of-words BoW introduced in computer vision by~\cite{Csurka04}, where each frame is represented by a histogram of features and the histogram bins are identified with the visual words of a visual dictionary. Stop lists and/or weighting schemes increase their accuracy~\cite{Sivic09}, since the former keeps very common patterns from contaminating the results by dropping the most common visual words, while the latter weights the contribution of each histogram element and transforms histograms into simple vectors. We adopt such a framework here and we build a $K$-length visual dictionary based on space-time interest points (STIP) \cite{LaptevMSR08}. The volume around such interest points is described by a 162-dimensional descriptor (HOG/HOF) by collecting image gradients and local optical flow. Moreover, we enable the \emph{inverse-document-frequency} (IDF) weighting scheme which down-weights words that occur frequently and works as a stop list as well~\cite{Manning08}. While the Hollywood-1 and Hollywood-2 datasets provide training and testing examples, a leave-one-out approach was followed for both the CONTACT and RAVEL datasets. 
\begin{figure*}[t!]
\centering
\begin{tabular}{|c|c|c|c|c|}
\hline
$ \tilde{Y}^{\text{AnsPhone}} $ & $X_1^{\text{AnsPhone}}$ & $ X_2^{\text{AnsPhone}}$ & $X_3^{\text{AnsPhone}} $ & $X_4^{\text{AnsPhone}} $ \\ \hline
\hline
$\tilde{y}_{20}$&
\includegraphics[width=0.19\textwidth]{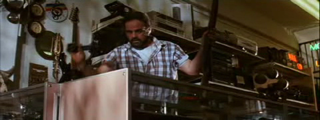} &
\includegraphics[width=0.19\textwidth]{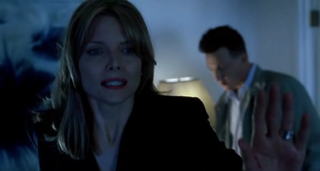} &
\includegraphics[width=0.19\textwidth]{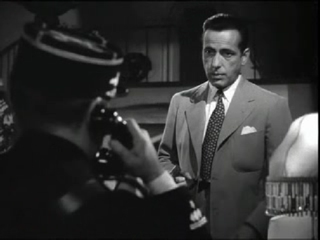} &
\includegraphics[width=0.19\textwidth]{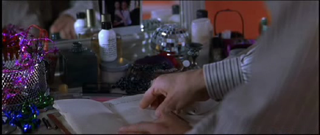} \\ \hline
$\tilde{y}_{60}$ &
\includegraphics[width=0.19\textwidth]{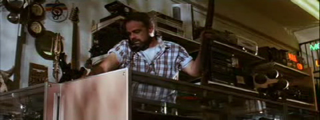} &
\includegraphics[width=0.19\textwidth]{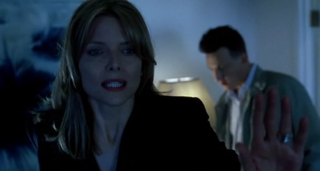} &
\includegraphics[width=0.19\textwidth]{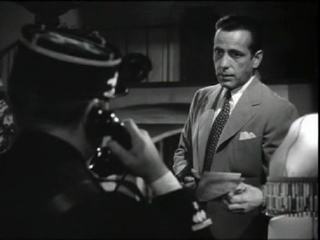} &
\includegraphics[width=0.19\textwidth]{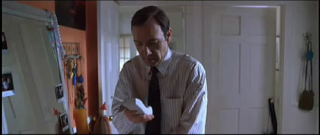}\\ \hline
$\tilde{y}_{90}$ &
\includegraphics[width=0.19\textwidth]{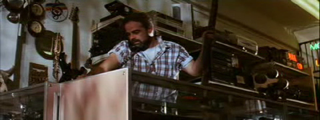} &
\includegraphics[width=0.19\textwidth]{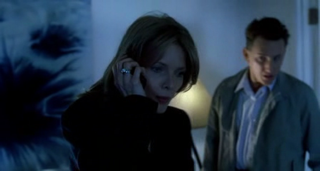} &
\includegraphics[width=0.19\textwidth]{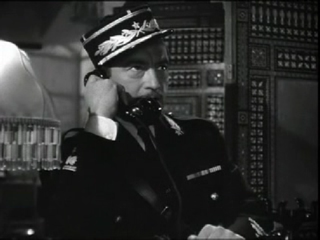} &
\includegraphics[width=0.19\textwidth]{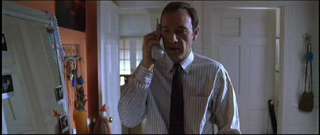}\\ \hline
$\tilde{y}_{120}$ &
\includegraphics[width=0.19\textwidth]{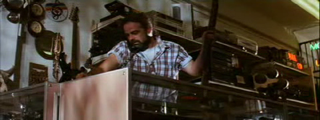} &
\includegraphics[width=0.19\textwidth]{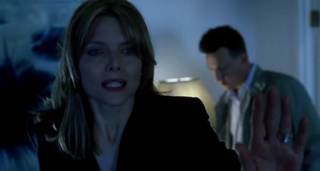} &
\includegraphics[width=0.19\textwidth]{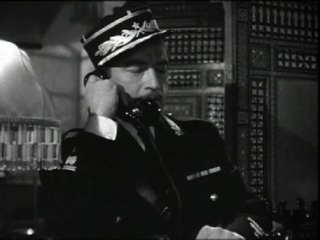} &
\includegraphics[width=0.19\textwidth]{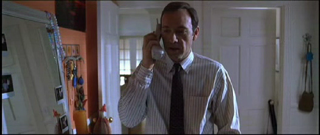} \\ \hline
\end{tabular}
\caption{\label{fig:meta-window}  Four metaframes sampled from the Answer-Phone category, shown horizontally. Four training examples of sequences associated with this category are shown vertically. As it can be seen, there is a large variability in the training dataset. The proposed test-frame-to-metaframe distance copes with these large within-category variations. }
\end{figure*}

\begin{figure*}[t!]
\begin{center}
\begin{tabular}{ccc}
\includegraphics[width=0.30\linewidth]{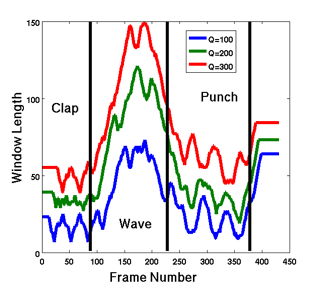} 
&
\includegraphics[width=0.30\linewidth]{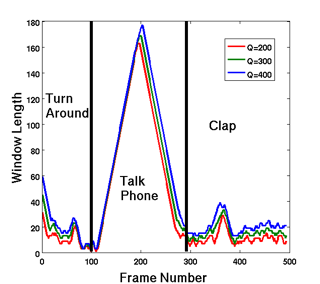} 
&
\includegraphics[width=0.30\linewidth]{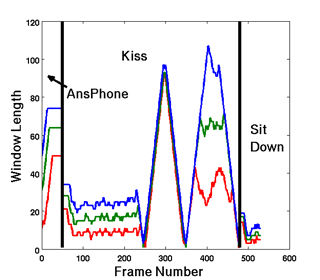}\\
(a) CONTACT 
&
(b) RAVEL 
&
(c) Hollywood-1\\
\end{tabular}
\end{center}
\caption{\label{fig:AdaptiveWindowChange} Size of the adaptive window at each frame of a video. One easily observes that frames located at action boundaries carry between-action information. Also for very large $Q$, the size of the adaptive window can be larger than the size of the action itself. This causes a drop in the recognition accuracy as $Q$ increases in the RAVEL (b) and Hollywood-1 (c) datasets.}
\end{figure*}

Since STIP features are used, which are sparse spatiotemporal features, we may evidently end up with all-zero feature vectors for frames with no motion. A symmetric temporal window around each frame can deal with this. A window of fixed size $W$ around each frame in the video never guarantees the overlap with frames of the non-static components of actions. Furthermore, it tends to over smooth self-discriminative frames for fast actors. Therefore, we use an adaptive-size window, namely we start from a single frame and we symmetrically grow the window size until a predefined number $Q$ of features  is gathered. This also offers a solution for variabilities in speed of different actors.

Fig.~\ref{fig:AdaptiveWindowChange} plots the adaptive window size as a function of time for the wave, talk-on-the-phone, and kiss action in the CONTACT, RAVEL, and Hollywood-1 datasets, respectively. 
The wave action in the CONTACT dataset consists of an actor starting from a standing position, raising hands to wave by just moving his fingers and coming back to a standing position. 
We can observe that a larger window size is required to aggregate $Q$ features in regions with almost no motions, while a smaller window size is  needed around regions containing large motions. 
A similar behavior is observed for the talk-on-the-phone action in the RAVEL dataset. Extremely large window sizes are required to aggregate a sufficient number of features, especially in the middle of the talk-on-the-phone action, since features from both the starting part, where the actor removes the phone from his pocket, and the ending part of the action, where the actor inserts the phone back in his pocket, have to be included. The same goes for the kiss action of the Hollywood-1 dataset. The fixed value of $Q$ reflects narrow windows when the persons move forward to kiss and wide windows when the persons are back to a stand position. In all plots, the action boundaries are marked with the vertical black lines. It is important to note that when we aggregate features across the action boundaries, the motion at the end of the previous action and the beginning of the next action is jointly encoded in the representation of this area. Experimentally, this shows that frames carrying between-category information act as an implicit weighting on the between-action transitions in our proposed algorithm.

The accuracy of continuous recognition can be affected by frames within a metaframe which are not perfectly aligned. This is true especially in the case of the Hollywood-1 dataset, as observed in Fig.~\ref{fig:meta-window}. This misalignment can, of course, lead to test-frame-to-metaframe misalignments. A test-frame-to-metaframe alignment error can then be propagated in the within-action and between-action transitions, thus affecting the overall recognition accuracy. To compensate for these misalignments, each frame of the test sequence is matched with several metaframes described by a symmetric window. We denote this window by $W_{\text{meta}}$. Note that $W_{\text{meta}}$ summarizes successive meta\-frames of the same category only.

\subsection{Results}

\begin{figure*}[t!]
\centering
\begin{tabular}{ccc}
 \includegraphics[width=0.32\linewidth]{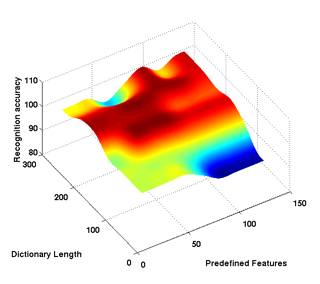} &
\includegraphics[width=0.32\linewidth]{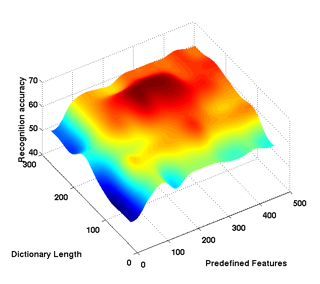}&
\includegraphics[width=0.32\linewidth]{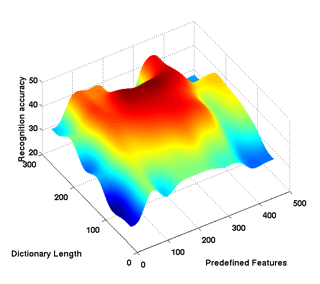}\\
(a) CONTACT &
(b) RAVEL &
(c) Hollywood-1 \\
\end{tabular}
\caption{\label{fig:VariationWRTQandK} A 3D plot of the recognition accuracy as a function of the dictionary length $K$ and of $Q$. An important observation is that the recognition accuracy remains constant with respect to $Q$ for the CONTACT dataset because the actions are always performed in the same order in the test data. In RAVEL and Hollywood-1 datasets the recognition accuracy first increases and then decreases with respect to $Q$ because for large values of $Q$ the per-frame feature representation is no longer discriminative.}
\end{figure*}

We discuss the performance of OP-DFW and the parameter value influence on the recognition accuracy. The proposed algorithm finds a string of actions that optimally aligns with a test sequence (\ref{eq:optimization-general}). As already discussed the predefined number of features $Q$ within an adaptive window is an important parameter, somehow correlated with the size $K$ of the feature vectors. We study how the choice of these parameters affect the recognition accuracy. The results are shown in Fig.~\ref{fig:VariationWRTQandK}. 
In the case of Hollywood-1 and Hollywood-2 datasets, we apply a cross-validation protocol to estimate the parameters $K$ and $Q$. Then these parameter values are used with all the test videos of these datasets. The CONTACT and RAVEL datasets do not contain enough training examples to perform cross-validation. Therefore we applied a leave-one-actor-out protocol and obtained values for the two parameters.
\color{black}

For the CONTACT dataset, the recognition accuracy remains constantly high with a varying $Q$, since the actions are performed in a predefined order. For the RAVEL and the Hollywood-1 datasets, the recognition accuracy first increases and then decreases with the value of $Q$,  as the actions are performed in a random order. Therefore, very large values of $Q$ make the frame representation no longer discriminative. There is a recognition performance peak when $Q$'s is in the range of $[50,150]$, $[200,350]$, and $[200,350]$ for the CONTACT, RAVEL and Hollywood-1 datasets, respectively, while $K\in[150,250]$ yields the maximizers with respect to the dictionary length for all datasets. As a consequence,  the recognition performance in RAVEL and Hollywood-1 are optimized for a similar range of both parameters $Q$ and $K$. 


We also study the impact of a varying $W_{\text{meta}}$ on the recognition accuracy. We observe that an increase in $W_{\text{meta}}$ leads to an increase in the recognition accuracy. This is because $W_{\text{meta}}$ accommodates for the variance observed in the frames within consecutive metaframes. This variance is illustrated in Fig.~\ref{fig:meta-window}. The recognition results are summarized in Table ~\ref{tab:MetaResults}. A larger $W_{\text{meta}}$ is required for the Hollywood-1 dataset as compared to the RAVEL dataset because of the imperfect alignment in the former and its higher variance of action duration (see \ref{fig:AvgVarDataSets}). As for the CONTACT dataset, changing the $W_{\text{meta}}$ does not influence the recognition accuracy because the alignment of frames within the metaframe is already close to perfect, as shown in Fig. ~\ref{fig:meta-frame}.

\begin{table}[t!]
\centering
\begin{tabular}{|c|c|c|c|c|c|}
\hline
$W_{\text{meta}}$ & 1 & 3 & 5 & 7 & 9\\ \hline
Hollywood-1 & 21.5 & 35.6 & 42.3 & 45.72 & 45.6\\ \hline
Ravel & 60.3 & 62.1 & 64.33 & 61.33 & 59.3\\ \hline
\end{tabular}
\caption{Results obtained with OP-DFW for different lengths of $W_{\text{meta}}$.}
\label{tab:MetaResults}
\end{table}

\begin{figure*}[t!]
\centering
\begin{tabular}{ccc}
\includegraphics[width=0.3\linewidth]{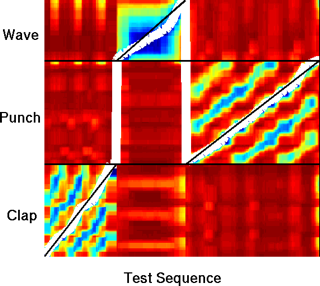} &
\includegraphics[width=0.3\linewidth]{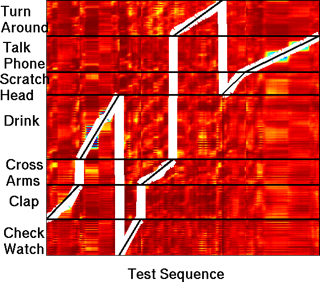}&
\includegraphics[width=0.3\linewidth]{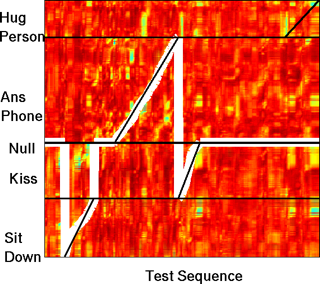}\\
\includegraphics[width=0.3\linewidth]{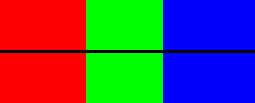} &
\includegraphics[width=0.3\linewidth]{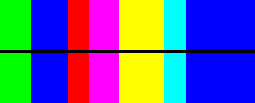}&
\includegraphics[width=0.3\linewidth]{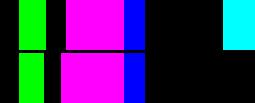}\\
(a) CONTACT &
(b) RAVEL &
(c) Hollywood-1 \\
\end{tabular}
\caption{\label{fig:DistMatPath} Illustration of the OP-DFW algorithm with three test sequences from the CONTACT, RAVEL, and Hollywood-1 datasets. The alignment path is shown in white. The diagonal segments of the path correspond to within-action transitions, vertical segments correspond to between-action transitions, and horizontal segments correspond to the null class. The bottom row compares the results of our method (lower stripes) with the ground truth. Black stripes correspond to the null-class.
}
\end{figure*}

\begin{figure*}[t!]
\centering
\begin{tabular}{ccc}
 \includegraphics[width=0.3\linewidth]{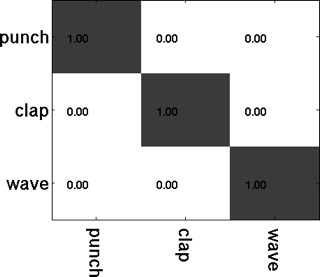} &
\includegraphics[width=0.3\linewidth]{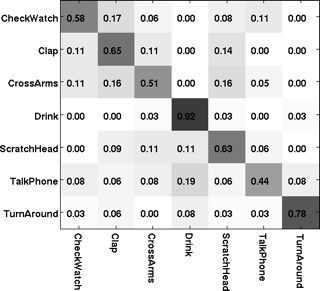}&
\includegraphics[width=0.3\linewidth]{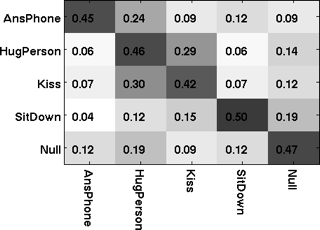}\\
(a) CONTACT &
(b) RAVEL &
(c) Hollywood-1 \\
\end{tabular}
\caption{\label{fig:ConfusionMats} Confusion matrices corresponding to the highest recognition accuracy obtained for each dataset. The diagonals indicate the percentage of frames correctly labelled in the test video sequences.}
\end{figure*}

The OP-DFW algorithm is illustrated on Fig.~\ref{fig:DistMatPath}. From this figure one can see that the alignment path found by our method (thick white line) follows well the ground truth (thin black line). Fig.~\ref{fig:ConfusionMats} shows the confusion matrices corresponding to the optimal parameters, $Q$ and $K$, for each dataset.
The method correctly distinguishes between relatively similar actions such as cross-arms and check-watch from RAVEL, but it has difficulties to discriminate between extremely similar actions such as kiss and hug-person from Hollywood-1. It is important to note that the talk-on-the-phone consistently yields low performance for all values of $Q$ and $K$. This is because it is an almost motionless action and hence a large window size is needed to gather a sufficient number of features (see also Fig.~\ref{fig:AdaptiveWindowChange}). However, a too large window size leads to a non-discriminative frame descriptor, as discussed above. 

\begin{figure}[t!]
\begin{center}
\includegraphics[width=0.95\linewidth]{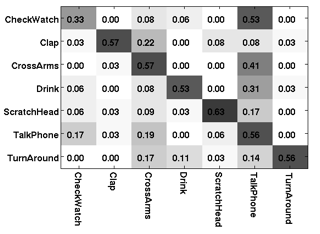}
\end{center}
\caption{\label{fig:ConfusionMatsFixed} Confusion matrix for fixed window size $W=15$ frames and $K=250$.}
\end{figure}

It is worth noting that the performance obtained with a fixed $Q$ (the number of spatiotemporal keypoints) is constantly better than the one obtained with a fixed $W$ (the temporal length of a symmetric window around each frame). We recall that a fixed $W$ implies an variable $Q$, and vice versa. There is however one exception, namely the talk-on-the-phone action just mentioned. The performance in this case yields better recognition performance with a fixed-length window. This is because fixing the length of $W$ prevents the window from being arbitrarily large, e.g., Fig.~\ref{fig:AdaptiveWindowChange}. A confusion matrix for $W=15$ and $K=250$ is shown on Fig.~\ref{fig:ConfusionMatsFixed}. Moreover, we experimentally verified that the \emph{idf} weighting scheme has no negative effects on the recognition accuracy and its impact is always beneficial. For example, the maximum recognition accuracy obtained with the \emph{idf} weighting is $45.72\%$ for the Hollywood-1 dataset, while the score is $43.9\%$ when we disable the weighting.

\subsection{Comparative Results}
\label{sec:comparisons}
In this section we compare our OP- and TP-DFW methods with \cite{HoaiLD11} and \cite{ShiWanCheSmoijcv11}, two state-of-the-art methods in continuous action recognition. 
While \cite{HoaiLD11} and \cite{ShiWanCheSmoijcv11} suggest two different action-level recognition methods -- based on multi-class support vector machines for the former and on semi-Markov models for the latter -- they share the same segmentation-level algorithm, which is essentially based on dynamic programming, namely the sequence-level procedure described in detail in section~\ref{ssec:TP-DFW}. Therefore, one can state that \cite{HoaiLD11} and \cite{ShiWanCheSmoijcv11} belong to the two-pass class of methods for simultaneous recognition and segmentation.

Matlab code of the methods described in  \cite{HoaiLD11} and \cite{ShiWanCheSmoijcv11} are publicly available. We use these authors' implementations in all our com\-pa\-ri\-sons. We were able to reproduce the results reported by the authors for the Hollywood-1 dataset. We also applied these two methods to the RAVEL dataset.
\color{black}

\begin{table}[t!]
\centering
\begin{tabular}{|c|c|c|}
\hline
Method  & Hollywood-1 & RAVEL\\ \hline\hline
\cite{ShiWanCheSmoijcv11} & 34.20 &55.40\\ \hline
\cite{HoaiLD11}  & 42.24 &59.90\\ \hline
One-pass DFW  & \textbf{45.72} & \textbf{64.33}\\ \hline
Two-pass DFW  & 42.12 & 59.70\\ \hline
One-pass HMM & 32.60 & 43.70\\ \hline
Two-pass HMM & 29.50 & 39.80\\ \hline
\end{tabular}
\caption{\label{tab:SummaryResults} Summary of results obtained with \textit{continuous recognition} methods using artificially merged  actions (Hollywood-1) and actions involving smooth transitions (RAVEL).}
\end{table}

\begin{table}[t!]
\begin{center}
\begin{tabular}{|c|c|c|}
\hline
Hollywood-1 & \cite{HoaiLD11} & OP-DFW \\ \hline\hline
Answer-Phone & 	35.0 & \textbf{45.0} \\ \hline
Hug-Person & 34.0 & \textbf{46.0} \\ \hline
Kiss      & \textbf{51.0} & 42.0 \\ \hline
Sit-Down & 45.0 & \textbf{50.0} \\ \hline
Null-Class & 47.0 & 47.0 \\ \hline
Average         &           42.2              &             \textbf{45.7} \\ \hline
\end{tabular}
\end{center}
\caption{\label{tab:hollywood1} Detailed \textit{continuous recognition} results obtained with the Hollywood-1 dataset.}
\end{table}

To make this comparison complete, we also implemented an HMM-based continuous recognition method.
A discriminative HMM, with a maximum mutual information (MMI) criteria, is trained with a predefined number of states.  We use the frame representation described in Sec.~\ref{ssec:implementation}. The training consists of two steps: firstly, an HMM is trained for each action category using the Baum-Welch algorithm; secondly, the mutual information between the training data and the action models is maximized \cite{gales2008application}. The number of states was set to $4$ and $6$ for the RAVEL and Hollywood-1 datasets, respectively. The HMM states are modeled with a Gaussian mixture. We experimentally found that $4$ components per state were sufficient to model the state's distribution. HMM-based sequence alignment can be used with both the one-pass and the two-pass schemes, where DTW is simply replaced by Viterbi. We refer to these two algorithms as OP-HMM and TP-HMM. We used a very recent toolkit to train the HMMs \cite{young2009htk}.

The results obtained with our two methods, with \cite{HoaiLD11,ShiWanCheSmoijcv11}, \cite{ShiWanCheSmoijcv11}, OP-HMM, and TP-HMM are summarized in Table~\ref{tab:SummaryResults}. The average precision for each action of the Hollywod-1 dataset, obtained with the proposed OP-DFW and with \cite{HoaiLD11,ShiWanCheSmoijcv11}, are summarized in Table~\ref{tab:hollywood1}.

\begin{figure*}[t!]
\centering
\includegraphics[width=0.120\linewidth]{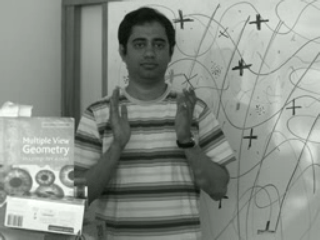}
\includegraphics[width=0.120\linewidth]{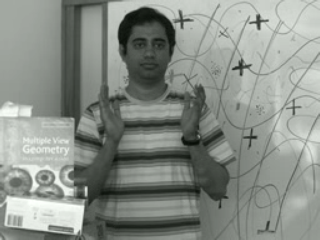}
\includegraphics[width=0.120\linewidth]{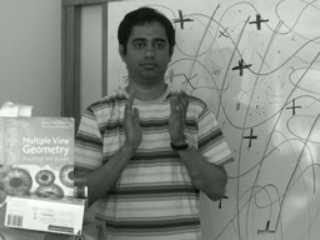}
\includegraphics[width=0.120\linewidth]{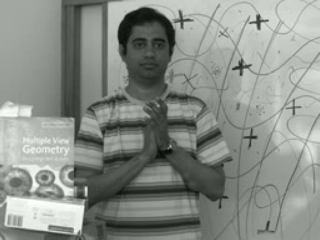}
\includegraphics[width=0.120\linewidth]{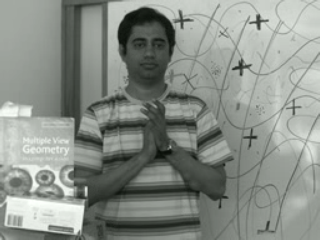}
\includegraphics[width=0.120\linewidth]{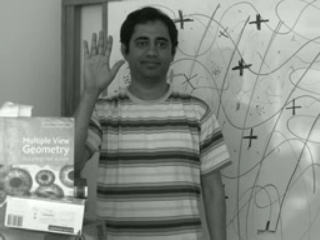}
\includegraphics[width=0.120\linewidth]{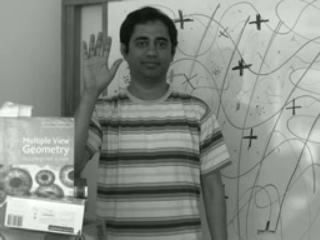}
\includegraphics[width=0.120\linewidth]{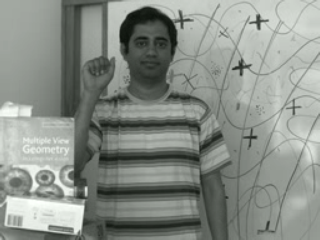}\\
\includegraphics[width=0.120\linewidth]{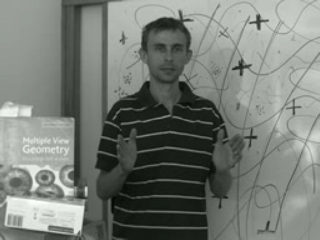}
\includegraphics[width=0.120\linewidth]{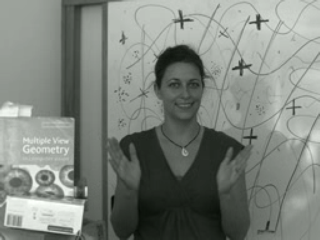}
\includegraphics[width=0.120\linewidth]{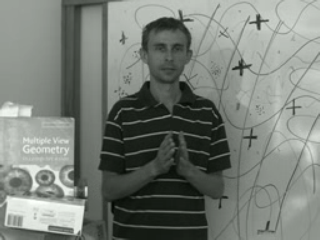}
\includegraphics[width=0.120\linewidth]{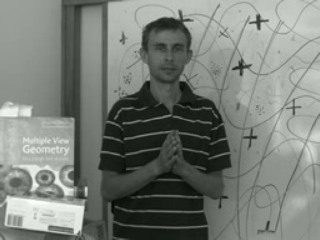}
\includegraphics[width=0.120\linewidth]{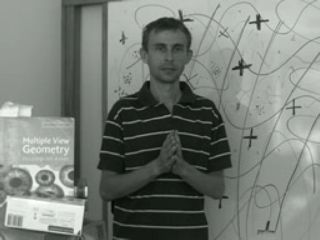}
\includegraphics[width=0.120\linewidth]{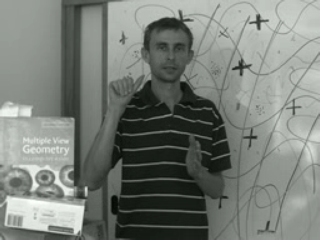}
\includegraphics[width=0.120\linewidth]{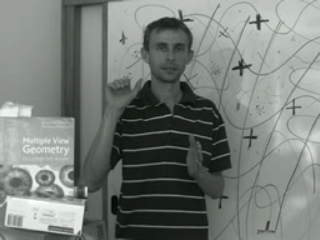}
\includegraphics[width=0.120\linewidth]{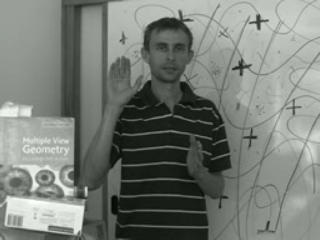}\\
(a) CONTACT (clap, smooth transition, wave) \\

\includegraphics[width=0.120\linewidth]{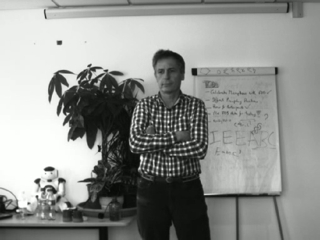}
\includegraphics[width=0.120\linewidth]{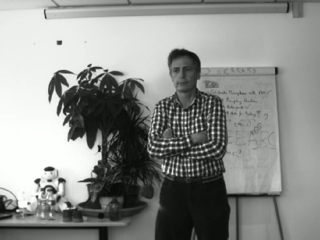}
\includegraphics[width=0.120\linewidth]{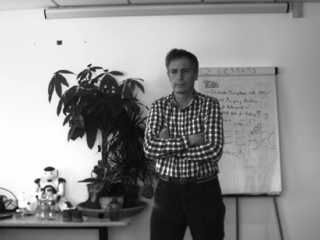}
\includegraphics[width=0.120\linewidth]{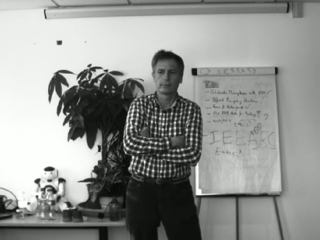}
\includegraphics[width=0.120\linewidth]{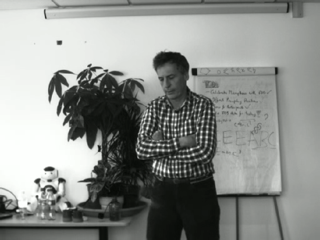}
\includegraphics[width=0.120\linewidth]{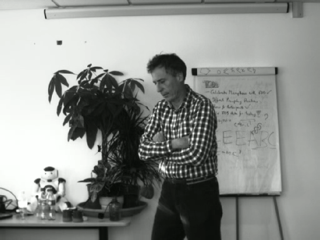}
\includegraphics[width=0.120\linewidth]{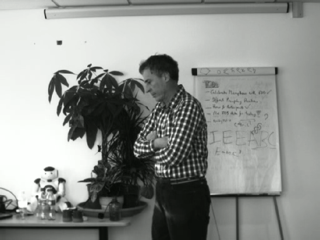}
\includegraphics[width=0.120\linewidth]{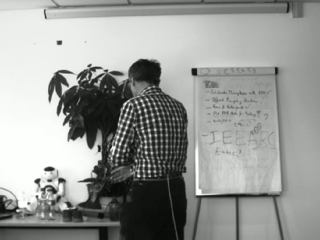}\\
\includegraphics[width=0.120\linewidth]{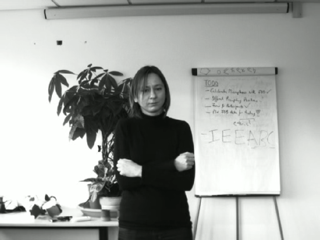}
\includegraphics[width=0.120\linewidth]{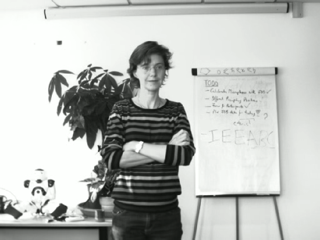}
\includegraphics[width=0.120\linewidth]{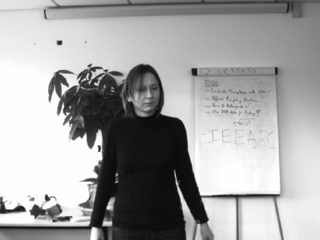}
\includegraphics[width=0.120\linewidth]{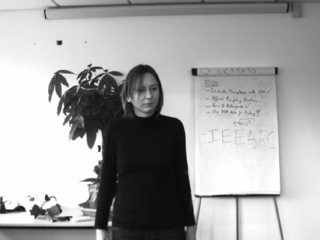}
\includegraphics[width=0.120\linewidth]{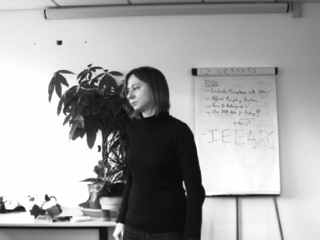}
\includegraphics[width=0.120\linewidth]{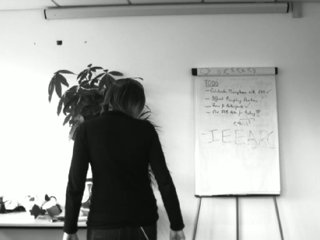}
\includegraphics[width=0.120\linewidth]{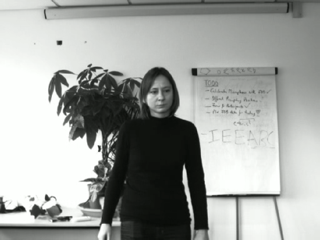}
\includegraphics[width=0.120\linewidth]{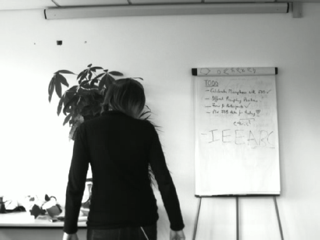}\\
(b) RAVEL (cross-arms, smooth transition, turn-around) \\
\includegraphics[width=0.120\linewidth]{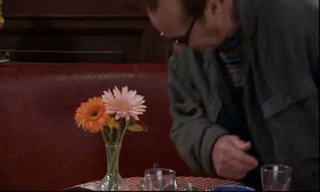}
\includegraphics[width=0.120\linewidth]{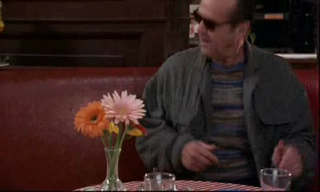}
\includegraphics[width=0.120\linewidth]{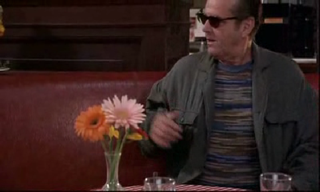}
\includegraphics[width=0.120\linewidth]{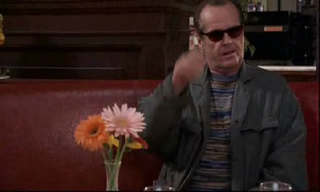}
\includegraphics[width=0.120\linewidth]{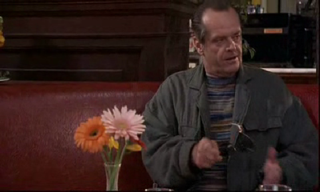}
\includegraphics[width=0.120\linewidth]{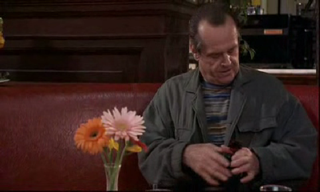}
\includegraphics[width=0.120\linewidth]{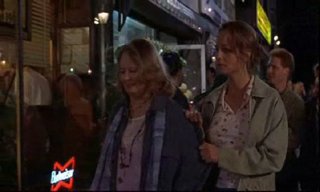}
\includegraphics[width=0.120\linewidth]{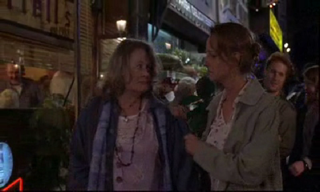}\\
\includegraphics[width=0.120\linewidth]{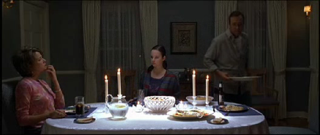}
\includegraphics[width=0.120\linewidth]{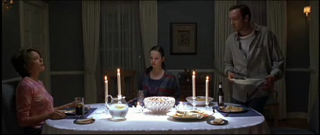}
\includegraphics[width=0.120\linewidth]{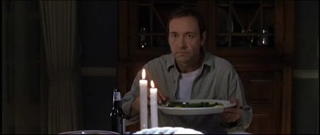}
\includegraphics[width=0.120\linewidth]{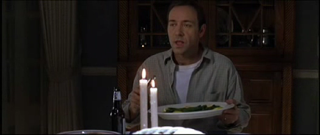}
\includegraphics[width=0.120\linewidth]{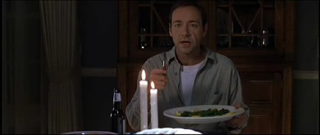}
\includegraphics[width=0.120\linewidth]{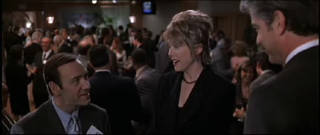}
\includegraphics[width=0.120\linewidth]{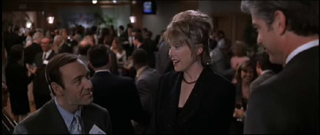}
\includegraphics[width=0.120\linewidth]{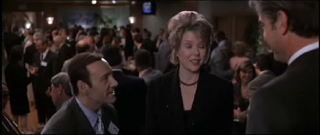}\\
(c) Hollywood-1 (sit-down and kiss) \\
\caption{\label{fig:OPTPDifference} Continuous recognition alignments between test data (top row) and training data (bottom row) obtained with OP-DFW.
In CONTACT and RAVEL the test frames are consistently aligned with training frames associated with the same actor. In Hollywood-1 some misalignments occur due to the complexity of the scenes.}
\end{figure*}

We notice that OP-DFW outperforms the other methods and that TP-DFW and \cite{HoaiLD11} yield comparative performance. The OP-DFW algorithm is able to exploit between-action information available in the feature vectors. Fig.~\ref{fig:OPTPDifference} shows that OP-DFW tends to align test-frames with example frames corresponding to the same actor in CONTACT and RAVEL. The two-pass methods cannot exploit between-action transition information embedded in the data. The performance obtained with the HMM-based algorithms is poor because, as already discussed, it is not possible to encode sub-units of actions. We conclude that the proposed OP-DFW method is better suited for continuous recognition than all the other two-pass methods, namely, TP-DFW, TP-HMM, \cite{HoaiLD11,ShiWanCheSmoijcv11} and \cite{ShiWanCheSmoijcv11}.

\begin{table}
\begin{center}
\begin{tabular}{|c|c|c|}
\hline
Hollywood-1 & \cite{LaptevMSR08} & DFW \\ \hline\hline
Answer-Phone  &          32.1           &                     \textbf{61.2} \\ \hline
Get-Out-Car      &    41.5        &         \textbf{58.3}           \\ \hline
Hand-Shake               &            32.3             &                    \textbf{55.8} \\ \hline
Hug-Person    &   40.6    &\textbf{57.7}                               \\ \hline
Sit-Down      &          38.6 &  \textbf{64.5}                                   \\ \hline
Sit-Up           &         18.2               &                        \textbf{59.0} \\ \hline
Stand-Up      &           50.5             &                        \textbf{66.7} \\ \hline
Kiss           &              53.3              &                     \textbf{56.3} \\ \hline
\textit{Average}         &    38.20      & \textbf{59.9}                                 \\ \hline
\end{tabular}
\caption{\label{tab:hollywood1-iso} Detailed \textit{isolated recognition} results obtained with the Hollywood-1 dataset.}
\end{center}
\end{table}

\begin{table}[h!]
\begin{center}
\begin{tabular}{|c|c|c|c|}
\hline
Hollywood-2 & \cite{ullah2010improving} & DFW & OP-DFW\\ \hline\hline
Answer-Phone  &     15.7                &                     \textbf{18.0}&    \\ \hline
Drive-Car      &         \textbf{87.6}                 &                    86.1&  \\ \hline
Eat               &         54.8                 &                    \textbf{56.3}&   \\ \hline
Fight-Person    &       \textbf{73.9}               &                      51.2&  56.4 \\ \hline
Get-Out-Car    &        33.4             &                        33.4&   \\ \hline
Hand-Shake    &        20.0               &                      \textbf{23.5}&   \\ \hline
Hug-Person    &         37.8                &                     \textbf{41.7}&   \\ \hline
Kiss           &             52.1               &                     \textbf{55.4}&   \\ \hline
Run            &             \textbf{71.1}               &              61.2&  65.6 \\ \hline
Sit-Down      &            \textbf{59.0}             &                       58.6&   \\ \hline
Sit-Up           &           23.9             &                        \textbf{31.0}&   \\ \hline
Stand-Up      &            53.3            &                        \textbf{54.4}&  \\ \hline
\textit{Average}         &           \textbf{48.6}              &                     47.6&   \\ \hline
\end{tabular}
\caption{\label{tab:hollywood2} Detailed \textit{isolated recognition} results obtained with the Hollywood-2 dataset.}
\end{center}
\end{table}

\addnote[isolated]{1}{While the primary objective of this paper is to address continuous action recognition, we also evaluate our algorithms in the case of isolated action recognition, i.e., the action boundaries are known and the task is reduced to action classification. Our approach offers two possibilities in this case, namely DFW and OP-DFW. The first possibility (DFW) corresponds to the combination of dynamic time warping (section~\ref{section:baseline_dtw}) with the frame-to-metaframe distance (sections~\ref{section:class_templates} and~\ref{section:dfw}). The second possibility (OP-DFW) consists in applying the algorithm of section~\ref{sec:op-dfw} to a test sequence that contains a single (isolated) action. In both cases we use a \textit{per-frame} representation based on STIP features and on HOG/HOF local descriptors, i.e., section~\ref{ssec:implementation}, hence a video is represented by a time series of feature vectors, one vector describing a window centered on each one of the frames in the video. The results obtained with DFW are compared with  \cite{LaptevMSR08} (Hollywood-1 dataset) with \cite{SCH12a} (RAVEL dataset), and with \cite{ullah2010improving} (Hollywood-2 dataset), while the results of OP-DFW are compared with \cite{ullah2010improving} (Holly\-wood-2 dataset). 
We chose these methods for our comparisons for two reasons: because they also use STIP features and because they provide recognition scores for each action category. Nevertheless, both \cite{LaptevMSR08} and \cite{ullah2010improving} make use of \textit{per-video} (global) representations, namely there is one feature vector associated with a video. It is interesting to notice that the best performing isolated action recognition methods also use global video representations, e.g., \cite{jiang2012trajectory}, \cite{solmaz2013classifying}, \cite{jain2013better}, and \cite{Wang2013action}. The latter is based on dense trajectory features and yields the best results on various datasets.}

\addnote[periodic-results]{1}{The isolated recognition results are detailed in table~\ref{tab:hollywood1-iso} (Hollywood-1) and table~\ref{tab:hollywood2} (Hollywood-2), and summarized in table~\ref{tab:IsolatedResults}.
One may notice that the proposed DFW method performs well on the Hollywood-1 dataset in comparison with \cite{LaptevMSR08}, while the performance on the Hollywood-2 dataset is less good. The reason is twofold:
\begin{itemize}
\item Firstly, the action boundaries are not very accurately annotated in the sequences of Holly\-wood-2. This means that the first and last frames of many sequences do not exactly correspond to the first and last frames of an action. Clearly, global-video representations perform better in such cases.
\item Secondly, Hollywood-2 contains two periodic actions, Run and Fight-Person and DFW cannot deal with actions composed of concatenated patterns. As it can be seen in table~\ref{tab:hollywood2}, this limitation is the main cause of degraded performance on this dataset.
\end{itemize}}

\begin{figure*}[tbh]
\centering
\begin{tabular}{cc}
\includegraphics[width=0.4\linewidth]{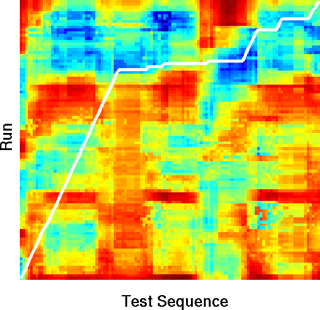} &
\includegraphics[width=0.4\linewidth]{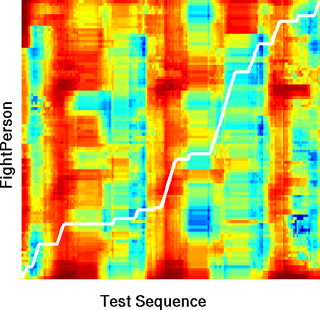}\\\vspace{0.8mm}
(a) Failure of DFW  (Run) & (b) Failure of DFW (Fight-Person)\\ 
\includegraphics[width=0.4\linewidth]{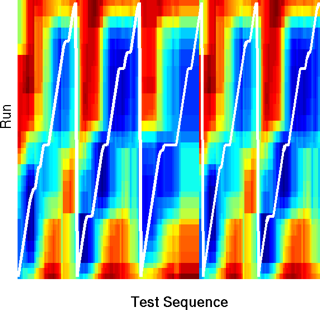} &
\includegraphics[width=0.4\linewidth]{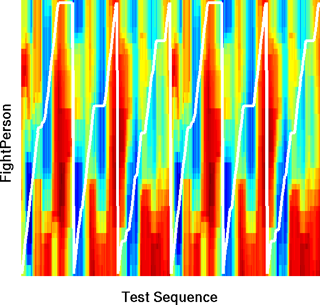}\\
(c) Correct alignment with OP-DFW (Run) & (d) Correct alignment with OP-DFW (Fight-Person)
\end{tabular}
\caption{\label{fig:Bad-Alignments} This figure shows the test-frame-to-metaframe distance grids (warm colors correspond to large discrepancies) and alignment paths (white lines) for the Run and Fight-Person sequences of Hollywood-2. The top plots, a) and b), show the result of applying DFW. In this case, the algorithm failed to properly align the test sequences with the corresponding class templates. One may notice that there is no obvious ``low-cost" path visible on these grids, therefore DFW fails to find a good alignment and a satisfactory score. The bottom plots, c) and d), show the result of applying OP-DFW to exactly the same test sequences, where Run and Fight-Person actions are modeled by their constituting motion patterns. Hence, OP-DFW jumps from one motion pattern to the next one. Examples of alignments with DFW and OP-DFW are shown on Fig.~\ref{fig:Failure-Alignments}.}
\end{figure*}

\begin{figure*}
\centering
\includegraphics[width=0.120\linewidth]{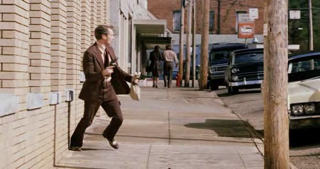}
\includegraphics[width=0.120\linewidth]{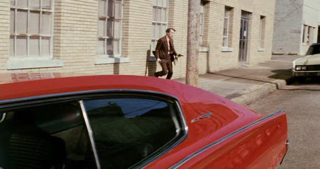}
\includegraphics[width=0.120\linewidth]{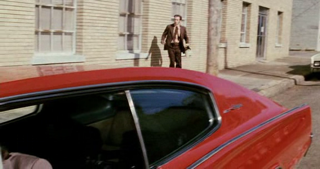}
\includegraphics[width=0.120\linewidth]{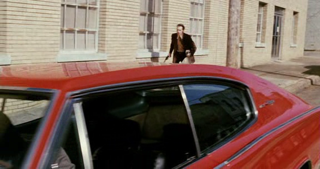}
\includegraphics[width=0.120\linewidth]{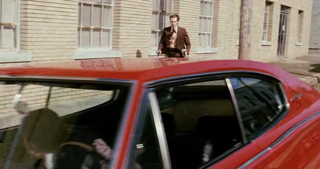}
\includegraphics[width=0.120\linewidth]{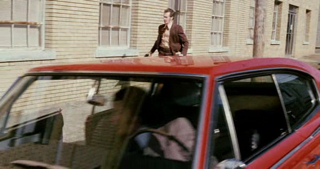}
\includegraphics[width=0.120\linewidth]{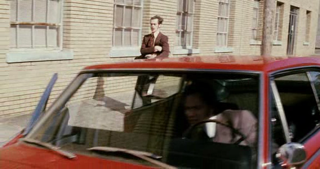}
\includegraphics[width=0.120\linewidth]{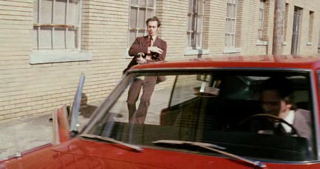}\\
(a) An example of a ``Run" test-sequence.\\
\includegraphics[width=0.120\linewidth]{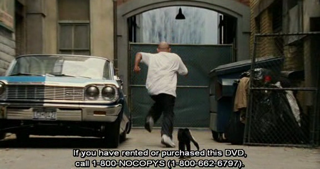}
\includegraphics[width=0.120\linewidth]{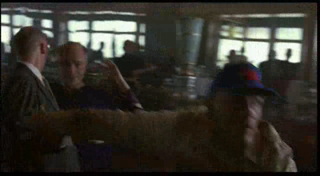}
\includegraphics[width=0.120\linewidth]{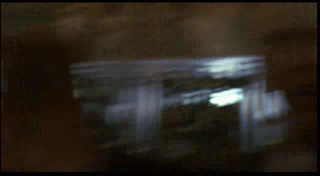}
\includegraphics[width=0.120\linewidth]{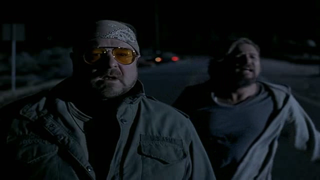}
\includegraphics[width=0.120\linewidth]{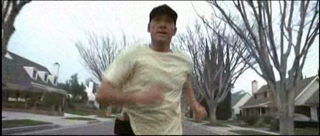}
\includegraphics[width=0.120\linewidth]{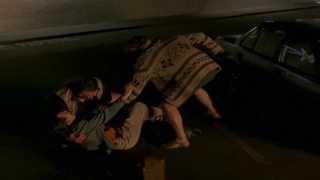}
\includegraphics[width=0.120\linewidth]{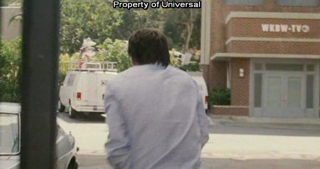}
\includegraphics[width=0.120\linewidth]{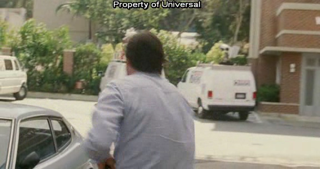}\\
(b) DFW failed to align this test sequence with metaframes from the ``Run" template.\\
\includegraphics[width=0.120\linewidth]{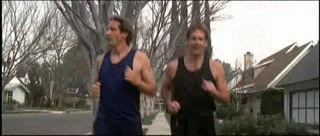}
\includegraphics[width=0.120\linewidth]{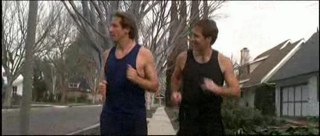}
\includegraphics[width=0.120\linewidth]{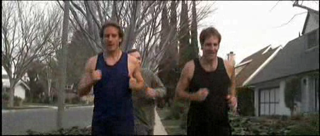}
\includegraphics[width=0.120\linewidth]{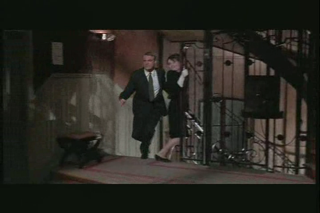}
\includegraphics[width=0.120\linewidth]{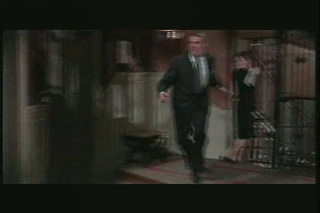}
\includegraphics[width=0.120\linewidth]{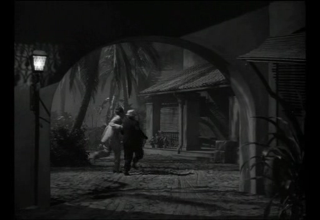}
\includegraphics[width=0.120\linewidth]{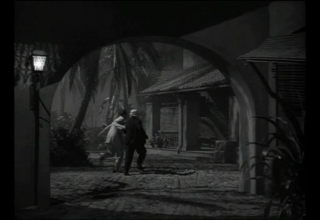}
\includegraphics[width=0.120\linewidth]{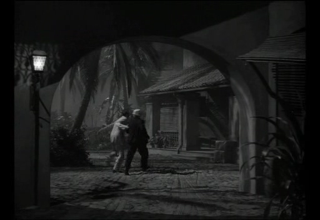}\\
(c) OP-DFW correctly aligned the test sequence with metaframes from the ``Run" template.\\
\includegraphics[width=0.120\linewidth]{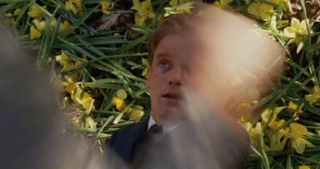}
\includegraphics[width=0.120\linewidth]{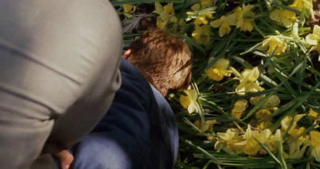}
\includegraphics[width=0.120\linewidth]{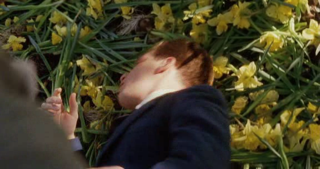}
\includegraphics[width=0.120\linewidth]{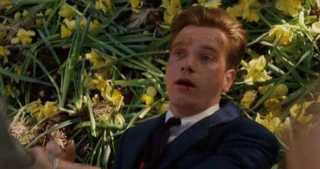}
\includegraphics[width=0.120\linewidth]{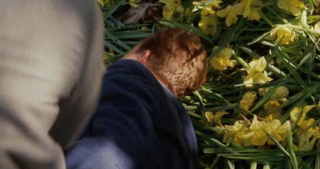}
\includegraphics[width=0.120\linewidth]{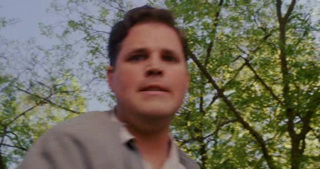}
\includegraphics[width=0.120\linewidth]{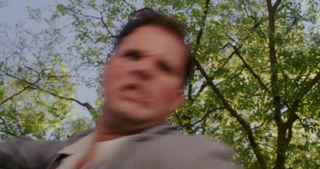}
\includegraphics[width=0.120\linewidth]{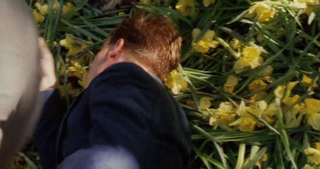}\\
(d) An example of a ``Fight-Person" test-sequence.\\
\includegraphics[width=0.120\linewidth]{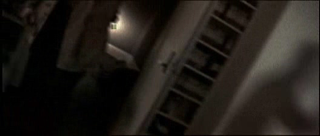}
\includegraphics[width=0.120\linewidth]{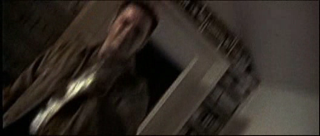}
\includegraphics[width=0.120\linewidth]{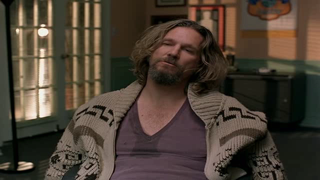}
\includegraphics[width=0.120\linewidth]{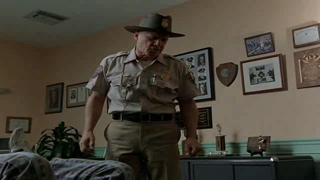}
\includegraphics[width=0.120\linewidth]{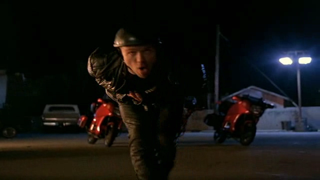}
\includegraphics[width=0.120\linewidth]{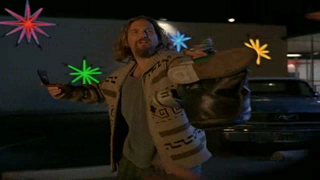}
\includegraphics[width=0.120\linewidth]{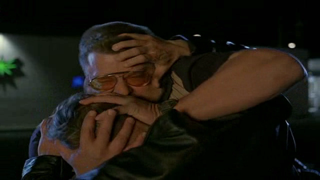}
\includegraphics[width=0.120\linewidth]{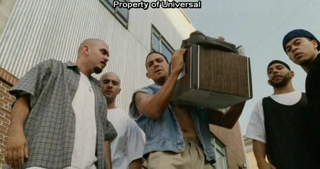}\\
(e) DFW failed to align this test sequence with metaframes from the ``Fight-Person" template.\\
\includegraphics[width=0.120\linewidth]{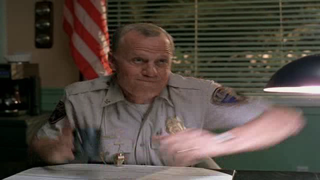}
\includegraphics[width=0.120\linewidth]{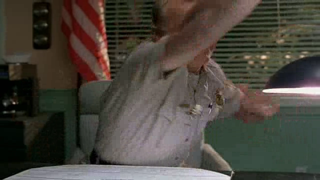}
\includegraphics[width=0.120\linewidth]{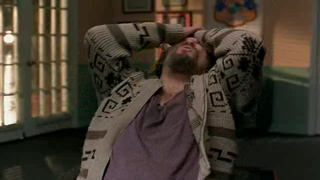}
\includegraphics[width=0.120\linewidth]{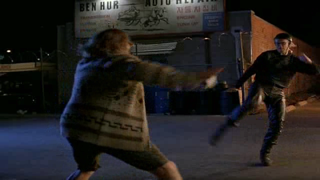}
\includegraphics[width=0.120\linewidth]{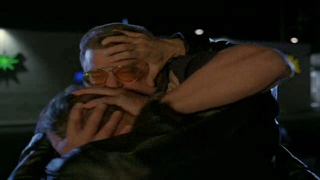}
\includegraphics[width=0.120\linewidth]{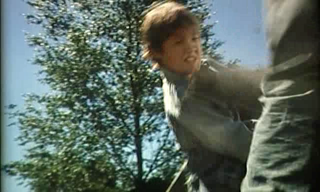}
\includegraphics[width=0.120\linewidth]{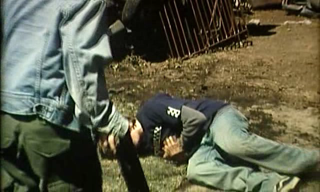}
\includegraphics[width=0.120\linewidth]{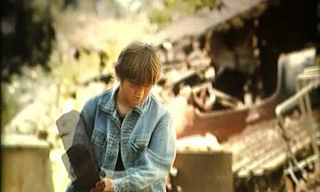}\\
(f) OP-DFW correctly aligned the test sequence with metaframes from the ``Fight-Person" template.\\
\caption{\label{fig:Failure-Alignments} Isolated recognition alignments using DFW and OP-DFW for periodic actions. The figure shows the alignments between a test video, (a) and (d), and metaframes from the training data for the Run and Fight-Person examples shown in Fig.~\ref{fig:Bad-Alignments}.}
\end{figure*}

\addnote[truncated]{1}{
We applied the OP-DFW method to the Fight-Person and Run examples of Holly\-wood-2 dataset (last column of table \ref{tab:hollywood2}).
 We modeled these two periodic actions as sequences of motion patterns (see section~\ref{sec:periodic-actions}); the alignment path generated by OP-DFW jumps an arbitrary number of times from the last frame of a pattern to the first frame of the next pattern. The difference between the alignment paths obtained with DFW and OP-DFW in the case of periodic actions are illustrated with two examples on figure~\ref{fig:Bad-Alignments}. A bad sequence alignment resulting from DFW is shown on Fig.~\ref{fig:Failure-Alignments}. }
\begin{table*}
\begin{center}
\begin{tabular}{|c|c|c|c|}
\hline
Method  & Hollywood-1 & RAVEL & Hollywood-2 \\ \hline\hline
\cite{LaptevMSR08} & 38.2 &- & - \\ \hline
\cite{SCH12a}  & - &68.4 & - \\ \hline
\cite{ullah2010improving} & - & - & 48.6 \\ \hline
\cite{Wang2013action} &    -    &  -  &            \textbf{64.3}         \\ \hline
Dynamic frame warping (DFW)  & \textbf{59.9} & \textbf{72.3} & 47.6 \\ \hline
\end{tabular}
\caption{Summary of \textit{isolated recognition} results for Hollywood-1, RAVEL, and Hollywood-2 datasets.}
\label{tab:IsolatedResults}
\end{center}
\end{table*}

\addnote[more-discussion]{1}{
Table \ref{tab:IsolatedResults} clearly shows that \cite{Wang2013action} outperforms both our method and \cite{ullah2010improving}. Moreover, the results reported in \cite{Wang2013action} (table~4) with several datasets reveal that Hollywood-2 remains, to date, the most challenging dataset for isolated recognition. Finally, it should be noted that global-video representations perform much better that per-frame representations in the presence of truncated actions.
}

\section{Discussion and Conclusions}
\label{section:conclusions}
The problem addressed in this paper, continuous action recognition, is more challenging than isolated recognition because action classification and action-boundary detection must be simultaneously carried out. For this reason, per-video (or global-video) representations, wide\-ly used by isolated recognition methods, are not appropriate because these representations do not embed between-action information, which is needed to detect transitions from one action to another. 

We propose a novel methodology based on sequence alignment using a per-frame representation. We build on the well known dynamic time warping framework and devise a novel representation of actions based on templates. An action-template is a time-series of meta\-frames that are produced by mutual alignments between all the training examples associated with a category. This per-metaframe representation allows to account for the large variabilities associated with actions in different contexts and performed by various people. 

When different actions are continuously performed, action-transition information is implicitly encoded in the data. Therefore, it is not only necessary to describe within-action information but between-action information as well. In order to incorporate this kind of knowledge we use a sliding temporal window, centered at each frame, and whose length is adjusted onto the data, such that a minimum number of spatiotemporal features are included in a frame descriptor, regardless of the ``speed of action". Using training examples that consist in long sequences composed of several annotated actions, it is therefore possible to encode between-action information in the first and last frames of each action in the training data. 

The proposed one-pass dynamic frame warping me\-thod simultaneously labels the frames of a test sequence and detects jumps between consecutive actions. We claim that the dual use of within-action and between-action transitions can also handle periodic motions formed of repetitive motion patterns. In this case, a periodic motion is viewed as a sequence of motion patterns and the OP-DFW algorithm is able to jump from pattern to pattern. We also discuss how a null-class can be incorporated into this one-pass framework. 

For completeness, we also propose an alternative continuous recognition algorithm that proceeds sequentially: recognition followed by segmentation, namely two-pass dynamic frame warping. We analyze in detail the algorithmic complexity of both algorithms and conclude that OP-DFW has the same order of complexity as the standard DTW algorithm. Hence, the proposed extension is barely more complex than isolated recognition based on sequence alignment. TP-DFW is more complex because it has to consider a large number of sub-sequences of the test sequence. Nevertheless, one advantage of TP-DFW over OP-DFW is that it can allow consecutive actions to slightly overlap. \addnote[more-truncation]{1}{Another advantage of TP-DFW is that the first pass can use any isolated action recognition method and representation, including discriminative classification using global-video descriptors. Therefore, TP-DFW has the potential to deal with truncated actions, for which global descriptors are more robust. We did not explore these numerous possibilities of TP-DFW and leave it for future work.}

We conducted a thorough experimental comparison which provides evidence that OP-DFW outperforms two state of the art methods that were recently proposed \cite{HoaiLD11}, \cite{ShiWanCheSmoijcv11}. TP-DFW and the two methods just cited share similar recognition and segmentation  performances, which is not surprising at all since the three methods rely on the same two-pass strategy.

Although isolated recognition is not the primarily objective of our work, we also conducted experiments with two isolated-action datasets, Hollywood-1 and Holly\-wood-2, and we provided detailed comparisons with \cite{LaptevMSR08} and \cite{ullah2010improving}. The periodic actions of Hollywood-1, e.g., Hand-Shake, do not seem to affect our method which outperforms  \cite{LaptevMSR08}. Hollywood-2 contains twelve actions. There are two periodic actions, Fight-Person and Run, for which \cite{ullah2010improving} performs better than our method. With both these datasets we used the same isolated recognition algorithm, based on dynamic frame warping, and which only makes use of within-action transitions. 

\addnote[explain-better]{1}{Nevertheless, if a periodic action is treated as a sequence composed of repetitive motion patterns, the proposed OP-DFW method is well suited, because it allows transitions between motion patterns. This has been validated experimentally with the CONTACT dataset and with the Run and Fight-Person of Hollywood-2.
Finally, it should be noted that the best performing isolated recognition methods use global-video representations, which perform much better than per-frame representations, such as ours, in the presence of truncated actions.
}

In the future we plan to extend our method to audio-visual data. Indeed, human activities are often synchronized with speech or prosodic sounds. It is therefore tempting to define audio-visual descriptors that combine visual and acoustic features. A recent gesture recognition challenge provides a multimodal dataset (color, depth and audio) \cite{escalera2013multi}. It is interesting to note that out of 17 teams that submitted their results, only one team used dynamic programming for segmentation. The winning team proposed an HMM method for audio recognition and segmentation, and dynamic time warping for skeletal classification.

\color{black}

\bibliographystyle{spbasic}   

\begin{thebibliography}{50}
\providecommand{\natexlab}[1]{#1}
\providecommand{\url}[1]{{#1}}
\providecommand{\urlprefix}{URL }
\expandafter\ifx\csname urlstyle\endcsname\relax
  \providecommand{\doi}[1]{DOI~\discretionary{}{}{}#1}\else
  \providecommand{\doi}{DOI~\discretionary{}{}{}\begingroup
  \urlstyle{rm}\Url}\fi
\providecommand{\eprint}[2][]{\url{#2}}

\bibitem[{Alameda-Pineda et~al(2013)Alameda-Pineda, Sanchez-Riera, Wienke,
  Franc, Cech, Kulkarni, Deleforge, and Horaud}]{ASWFCKDH13}
Alameda-Pineda X, Sanchez-Riera J, Wienke J, Franc V, Cech J, Kulkarni K,
  Deleforge A, Horaud RP (2013) {RAVEL: A}n annotated corpus for training
  robots with audiovisual abilities. Journal on Multimodal User Interfaces
  7(1-2):79--91

\bibitem[{Alon et~al(2009)Alon, Athitsos, Yuan, and
  Sclaroff}]{AlonAthistsosYuanSclaroff2009}
Alon J, Athitsos V, Yuan Q, Sclaroff S (2009) A unified framework for gesture
  recognition and spatiotemporal gesture segmentation. IEEE Transactions on
  Pattern Analysis and Machine Intelligence 31(9):1685--1699

\bibitem[{Blackburn and Ribeiro(2007)}]{blackburn2007human}
Blackburn J, Ribeiro E (2007) Human motion recognition using isomap and dynamic
  time warping. In: Human Motion--Understanding, Modeling, Capture and
  Animation, Springer, pp 285--298

\bibitem[{Boyd and Vandenberghe(2004)}]{BoydBook}
Boyd S, Vandenberghe L (2004) Convex Optimization. Cambridge University Press,
  New York, NY, USA

\bibitem[{Brendel and Todorovic(2010)}]{BrendelT10}
Brendel W, Todorovic S (2010) Activities as time series of human postures. In:
  ECCV (2)

\bibitem[{Chen et~al(2001)Chen, Donoho, and Saunders}]{Chen2001}
Chen SS, Donoho DL, Saunders MA (2001) Atomic decomposition by basis pursuit.
  SIAM Rev 43(1):129--159

\bibitem[{Csurka et~al(2004)Csurka, Dance, Fan, Willamowski, and
  Bray}]{Csurka04}
Csurka G, Dance CR, Fan L, Willamowski J, Bray C (2004) Visual categorization
  with bags of keypoints. In: ECCV Workshop on Statistical Learning in Computer
  Vision

\bibitem[{Escalera et~al(2013)Escalera, Gonz{\`a}lez, Bar{\'o}, Reyes, Lopes,
  Guyon, Athitsos, and Escalante}]{escalera2013multi}
Escalera S, Gonz{\`a}lez J, Bar{\'o} X, Reyes M, Lopes O, Guyon I, Athitsos V,
  Escalante HJ (2013) Multi-modal gesture recognition challenge 2013: {D}ataset
  and results. In: ChaLearn Multi-modal Gesture Recognition Grand Challenge and
  Workshop, 15th ACM International Conference on Multimodal Interaction

\bibitem[{Evangelidis and Psarakis(2008)}]{Evangelidis08}
Evangelidis GD, Psarakis EZ (2008) Parametric image alignment using enhanced
  correlation coefficient maximization. IEEE Transactions on Pattern Analysis
  and Machine Intelligence 30(10):1858--1865

\bibitem[{Gales and Young(2008)}]{gales2008application}
Gales M, Young S (2008) The application of hidden {M}arkov models in speech
  recognition. Foundations and Trends in Signal Processing

\bibitem[{Gill et~al(2011)Gill, Wang, and Molnar}]{Gill2011}
Gill PR, Wang A, Molnar A (2011) The in-crowd algorithm for fast basis pursuit
  denoising. IEEE Transactions on Signal Processing 59(10):4595--4605

\bibitem[{Gong and Medioni(2011)}]{GongMedioni2011}
Gong D, Medioni G (2011) Dynamic manifold warping for view invariant action
  recognition. In: IEEE International Conference on Computer Vision, IEEE, pp
  571--578

\bibitem[{Hienz et~al(1999)Hienz, Bauer, and Kraiss}]{HienzBauerKraiss1999}
Hienz H, Bauer B, Kraiss KF (1999) {HMM}-based continuous sign language
  recognition using stochastic grammars. In: Braffort A, Gherbi R, Gibet S,
  Teil D, Richardson J (eds) Gesture-Based Communication in Human-Computer
  Interaction, Lecture Notes in Computer Science, vol 1739, Springer Berlin /
  Heidelberg, pp 185--196

\bibitem[{Hoai et~al(2011)Hoai, Lan, and De~la Torre}]{HoaiLD11}
Hoai M, Lan ZZ, De~la Torre F (2011) Joint segmentation and classification of
  human actions in video. In: CVPR

\bibitem[{Ikizler and Duygulu(2009)}]{Ikizler:2009}
Ikizler N, Duygulu P (2009) Histogram of oriented rectangles: A new pose
  descriptor for human action recognition. IVC 27

\bibitem[{Jain et~al(2013)Jain, J{\'e}gou, and Bouth{\'e}my}]{jain2013better}
Jain M, J{\'e}gou H, Bouth{\'e}my P (2013) Better exploiting motion for better
  action recognition. In: Computer Vision and Pattern Recognition, pp
  2555--2562

\bibitem[{Jiang et~al(2012)Jiang, Dai, Xue, Liu, and Ngo}]{jiang2012trajectory}
Jiang YG, Dai Q, Xue X, Liu W, Ngo CW (2012) Trajectory-based modeling of human
  actions with motion reference points. In: European Conference on Computer
  Vision, Springer, pp 425--438

\bibitem[{Kulkarni et~al(2008)Kulkarni, Cherla, Kale, and
  Ramasubramanian}]{kulkarniECCV2008}
Kulkarni K, Cherla S, Kale A, Ramasubramanian V (2008) A framework for indexing
  human actions in video. ECCV MLVMA Workshop

\bibitem[{Laptev et~al(2008)Laptev, Marszalek, Schmid, and
  Rozenfeld}]{LaptevMSR08}
Laptev I, Marszalek M, Schmid C, Rozenfeld B (2008) Learning realistic human
  actions from movies. In: CVPR

\bibitem[{Lee and Rabiner(1989)}]{lee1989frame}
Lee C, Rabiner L (1989) A frame-synchronous network search algorithm for
  connected word recognition. IEEE Transactions on Acoustics, Speech and Signal
  Processing 37(11):1649--1658

\bibitem[{Liang and Ouhyoung(1998)}]{liang1998real}
Liang R, Ouhyoung M (1998) A real-time continuous gesture recognition system
  for sign language. In: Third IEEE International Conference on Automatic Face
  and Gesture Recognition, 1998, IEEE, pp 558--567

\bibitem[{Lv and Nevatia(2006)}]{LvNevatia2006}
Lv F, Nevatia R (2006) Recognition and segmentation of 3-{d} human action using
  {HMM} and multi-class {AdaBoost}. European Conference on Computer Vision pp
  359--372

\bibitem[{Lv and Nevatia(2007)}]{LvNevatia2007}
Lv F, Nevatia R (2007) Single view human action recognition using key pose
  matching and {V}iterbi path searching. In: CVPR

\bibitem[{Manning et~al(2008)Manning, Raghavan, and Sch\"utze}]{Manning08}
Manning C, Raghavan P, Sch\"utze H (2008) Introduction to Information
  Retrieval. Cambridge University Press

\bibitem[{Marszalek et~al(2009)Marszalek, Laptev, and
  Schmid}]{marszalek2009actions}
Marszalek M, Laptev I, Schmid C (2009) Actions in context. In: IEEE Conference
  on Computer Vision and Pattern Recognition, IEEE, pp 2929--2936

\bibitem[{Morency et~al(2007)Morency, Quattoni, and
  Darrell}]{morency2007latent}
Morency L, Quattoni A, Darrell T (2007) Latent-dynamic discriminative models
  for continuous gesture recognition. In: Computer Vision and Pattern
  Recognition, IEEE, pp 1--8

\bibitem[{Mueller(2007)}]{Mueller2007}
Mueller M (2007) Dynamic time warping. In: Information Retrieval for Music and
  Motion, Springer Berlin Heidelberg, pp 69--84

\bibitem[{Ney(1984)}]{ney1984use}
Ney H (1984) The use of a one-stage dynamic programming algorithm for connected
  word recognition. IEEE Transactions on Acoustics, Speech and Signal
  Processing 32(2):263--271

\bibitem[{Ney and Ortmanns(1999)}]{NeyOrtmanns1999}
Ney H, Ortmanns S (1999) Dynamic programming search for continuous speech
  recognition. IEEE Signal Processing Magazine 16(5):64 --83

\bibitem[{Ning et~al(2008)Ning, Xu, Gong, and Huang}]{ning2008latent}
Ning H, Xu W, Gong Y, Huang T (2008) Latent pose estimator for continuous
  action recognition. In: European Conference on Computer Vision, Springer, pp
  419--433

\bibitem[{Rabiner and Juang(1993)}]{RabinerJuang}
Rabiner L, Juang B (1993) Fundamentals of Speech Recognition. Prentice Hall

\bibitem[{Sakoe(1979)}]{Sakoe1979}
Sakoe H (1979) Two-level {DP}-matching -- a dynamic programming-based pattern
  matching algorithm for connected word recognition. IEEE Transactions on
  Acoustic, Speech, and Signal Processing 27(6):588 -- 595

\bibitem[{Sanchez-Riera et~al(2012)Sanchez-Riera, Cech, and Horaud}]{SCH12a}
Sanchez-Riera J, Cech J, Horaud RP (2012) Action recognition robust to
  background clutter by using stereo vision. In: The Fourth International
  Workshop on Video Event Categorization, Tagging and Retrieval, Springer, LNCS

\bibitem[{Shi et~al(2011)Shi, Wang, Cheng, and Smola}]{ShiWanCheSmoijcv11}
Shi Q, Wang L, Cheng L, Smola A (2011) Discriminative human action segmentation
  and recognition using {SMM}s. IJCV 93(1):22--32

\bibitem[{Sigal et~al(2010)Sigal, Balan, and Black}]{sigal2010humaneva}
Sigal L, Balan A, Black M (2010) Humaneva: Synchronized video and motion
  capture dataset and baseline algorithm for evaluation of articulated human
  motion. International journal of computer vision 87(1):4--27

\bibitem[{Sivic and Zisserman(2009)}]{Sivic09}
Sivic J, Zisserman A (2009) Efficient visual search of videos cast as text
  retrieval. IEEE Trans on PAMI 31(4):591--606

\bibitem[{Sminchisescu et~al(2006)Sminchisescu, Kanaujia, and
  Metaxas}]{SminchicescuKanaujiaMetaxas2006}
Sminchisescu C, Kanaujia A, Metaxas DN (2006) Conditional models for contextual
  human motion recognition. CVIU 104(2-3):210--220

\bibitem[{Solmaz et~al(2013)Solmaz, Assari, and Shah}]{solmaz2013classifying}
Solmaz B, Assari SM, Shah M (2013) Classifying web videos using a global video
  descriptor. Machine vision and applications 24(7):1473--1485

\bibitem[{Starner et~al(1998)Starner, Weaver, and Pentland}]{starner1998real}
Starner T, Weaver J, Pentland A (1998) Real-time american sign language
  recognition using desk and wearable computer based video. IEEE Transactions
  on Pattern Analysis and Machine Intelligence 20(12):1371--1375

\bibitem[{Tropp and Gilbert(2007)}]{Trop2007}
Tropp JA, Gilbert AC (2007) Signal recovery from random measurements via
  orthogonal matching pursuit. IEEE Transactions on Information Theory
  53(12):4655--4666

\bibitem[{Ullah et~al(2010)Ullah, Parizi, and Laptev}]{ullah2010improving}
Ullah MM, Parizi SN, Laptev I (2010) Improving bag-of-features action
  recognition with non-local cues. In: British Machine Vision Conference

\bibitem[{Vail et~al(2007)Vail, Veloso, and Lafferty}]{vail2007conditional}
Vail D, Veloso M, Lafferty J (2007) Conditional random fields for activity
  recognition. In: Proceedings of the 6th international joint conference on
  Autonomous agents and multiagent systems, ACM, p 235

\bibitem[{Vintsyuk(1971)}]{vintsyuk1971element}
Vintsyuk T (1971) Element-wise recognition of continuous speech composed of
  words from a specified dictionary. Cybernetics and Systems Analysis
  7(2):361--372

\bibitem[{Vogler and Metaxas(1998)}]{vogler1998asl}
Vogler C, Metaxas D (1998) {ASL} recognition based on a coupling between {HMM}s
  and {3D} motion analysis. In: Sixth International Conference on Computer
  Vision, pp 363--369

\bibitem[{Vogler and Metaxas(2001)}]{vogler2001framework}
Vogler C, Metaxas D (2001) A framework for recognizing the simultaneous aspects
  of american sign language. Computer Vision and Image Understanding
  81(3):358--384

\bibitem[{Wang and Schmid(2013)}]{Wang2013action}
Wang H, Schmid C (2013) Action recognition with improved trajectories. In:
  International Conference on Computer Vision

\bibitem[{Young et~al(1989)Young, Russell, and Thornton}]{young1989token}
Young S, Russell NH, Thornton J (1989) Token passing: a simple conceptual model
  for connected speech recognition systems. Tech. Rep.~38, University of
  Cambridge, Department of Engineering

\bibitem[{Young et~al(1993)Young, Woodland, and Byrne}]{young1993htk}
Young S, Woodland P, Byrne W (1993) {HTK}: Hidden {Markov} model toolkit v1. 5.
  Tech. rep., University of Cambridge, Department of Engineering

\bibitem[{Young et~al(2009)Young, Evermann, Kershaw, Moore, Odell, Ollason,
  Valtchev, and Woodland}]{young2009htk}
Young S, Evermann G, Kershaw D, Moore G, Odell J, Ollason D, Valtchev V,
  Woodland P (2009) The {HTK} book. Tech. rep., University of Cambridge,
  Department of Engineering

\bibitem[{Zhou and la~Torre(2009)}]{ZhouT09}
Zhou F, la~Torre FD (2009) Canonical time warping for alignment of human
  behavior. In: NIPS

\end{thebibliography}

\end{document}